\documentclass[10pt,twocolumn,letterpaper]{article}

\usepackage[pagenumbers]{cvpr} % To force page numbers, e.g. for an arXiv version
\usepackage{times}
\usepackage{epsfig}
\usepackage{graphicx}
\usepackage{amsmath}
\usepackage{amssymb}
\usepackage{booktabs}
\usepackage{multirow}
\usepackage{float}
\usepackage{multicol}
\usepackage{lipsum} % for dummy text only
\usepackage{xargs}  % Use more than one optional parameter in a new commands
\usepackage{xcolor}  % Coloured text etc.

\usepackage[titletoc,title]{appendix}

\usepackage[pagebackref=true,breaklinks=true,colorlinks,bookmarks=false]{hyperref}
\usepackage[numbers,sort,compress]{natbib} % reference in increasing order
\usepackage[capitalize]{cleveref}
\crefname{section}{Sec.}{Secs.}
\Crefname{section}{Section}{Sections}
\Crefname{table}{Tab.}{Tabs.}
\crefname{table}{Tab.}{Tabs.}
\newcommand*{\newcite}[1]{~\cite{#1}}

\def\eg{e.g.,~}      % for example
\def\ie{i.e.,~}      % that is, in other words
\def\linemod{LineMOD\ } 

\newcommand{\comment}[1]{}
\newcommand\norm[1]{\left\lVert#1\right\rVert}

\def\cvprPaperID{5164} % *** Enter the CVPR Paper ID here
\def\confName{CVPR}
\def\confYear{2022}

\begin{document}
% include the main paper(8 pages)
%\include{latex/input}

%%%%%%%%% TITLE 
\title{GAML: Geometry-Aware Meta-Learner for Cross-Category 6D Pose Estimation}

\author{Yumeng Li$^{1}$\thanks{Equal contribution} \ \  Ning Gao$^{1,2}$\footnotemark[1] \  \ Hanna Ziesche$^{1}$ \ Gerhard Neumann$^{2}$\\
$^1$Bosch Center for Artificial Intelligence\ \  $^2$Autonomous Learning Robots, KIT\\
{\tt\small \{yumeng.li, ning.gao, hanna.ziesche\}@de.bosch.com \ {gerhard.neumann}@kit.edu}
}
\maketitle

%%%%%%%%% ABSTRACT
\begin{abstract}
We present a novel meta-learning approach for 6D pose estimation on unknown objects. In contrast to ``instance-level" and ``category-level" pose estimation methods, our algorithm learns object representation in a category-agnostic way, which endows it with strong generalization capabilities across object categories. Specifically, we employ a neural process-based meta-learning approach to train an encoder to capture texture and geometry of an object in a latent representation, based on very few RGB-D images and ground-truth keypoints. The latent representation is then used by a simultaneously meta-trained decoder to predict the 6D pose of the object in new images. Furthermore, we propose a novel geometry-aware decoder for the keypoint prediction using a Graph Neural Network (GNN), which explicitly takes geometric constraints specific to each object into consideration. 
To evaluate our algorithm, extensive experiments are conducted on the \linemod dataset, and on our new fully-annotated synthetic datasets generated from Multiple Categories in Multiple Scenes (MCMS). Experimental results demonstrate that our model performs well on unseen objects with very different shapes and appearances. Remarkably, our model also shows robust performance on occluded scenes although trained fully on data without occlusion. To our knowledge, this is the first work exploring \textbf{cross-category level} 6D pose estimation. 
\end{abstract}

%%%%%%%%% BODY TEXT
\section{Introduction}
Estimating the 6D pose of an object is of practical interest for many real-world applications such as robotic grasping, autonomous driving and augmented reality (AR). Prior work has investigated instance-level 6D pose estimation\newcite{DenseFusion,pvnet,pvn3d,ffb6d}, where the objects are predefined. Although achieving satisfying performance, these methods are prone to overfit to specific objects and thus suffer from poor generalization. Due to the high variety of objects with different colors and shapes in the real-world, it is impractical to retrain the model every time new objects come in, which is time-consuming and data inefficient.
\begin{figure}[t]
\centering
	\includegraphics[width=\columnwidth]{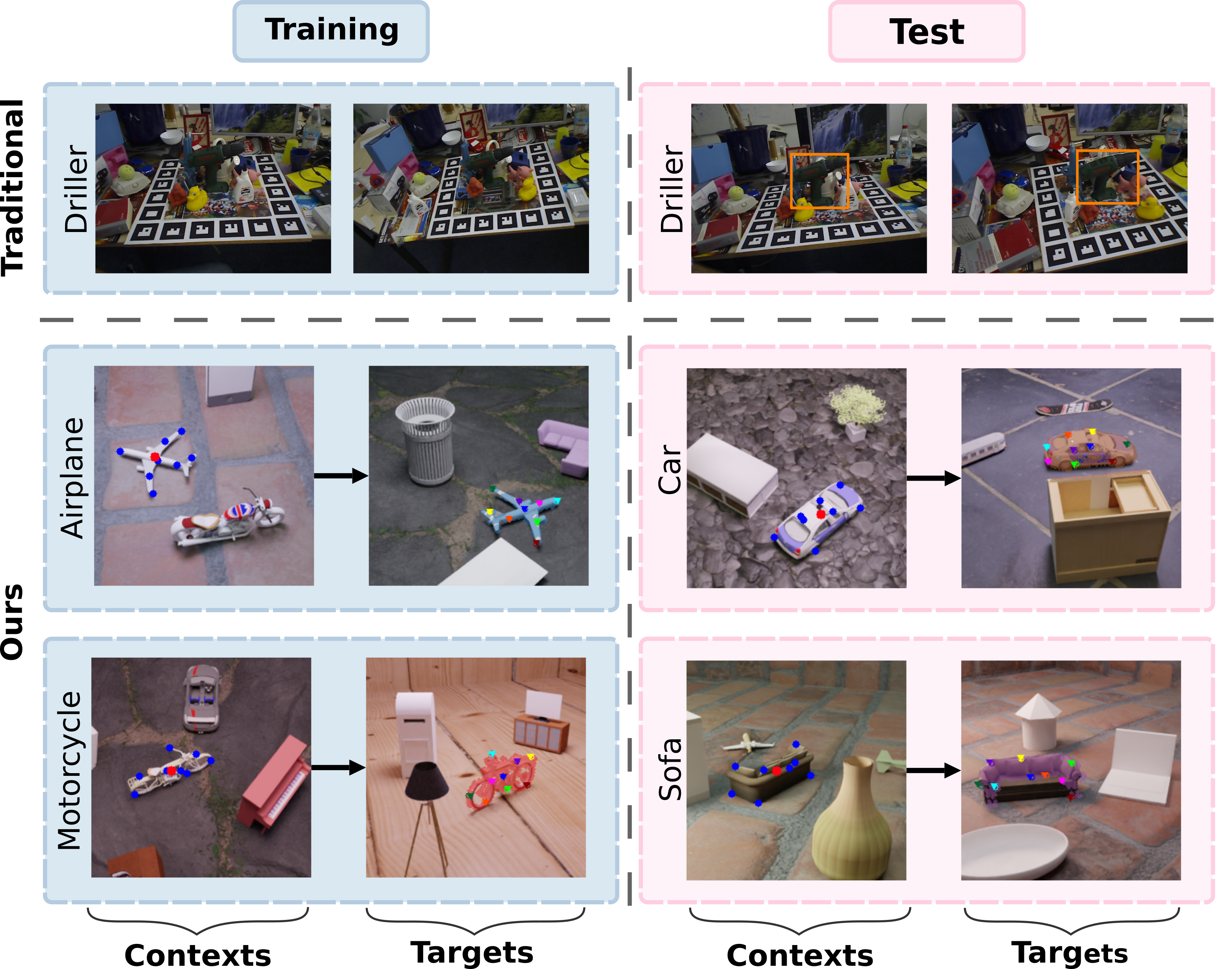}
		\caption{Illustration of the difference between traditional instance-level 6D pose estimation methods and our approach. Unlike other methods, our proposed approach generalizes to novel objects given a few context observations. The projected ground-truth keypoints are visualized as blue points in the context images. The predicted segmentation and keypoints are visualized in the target images.}
		\label{fig:intro}
\end{figure}
Recently, this issue has raised increasing attention in the community and several approaches\newcite{nocs,cass,6-pack,FS-Net,NeuralSynthesis,SGPA_2021_ICCV} have been proposed for category-level 6D pose estimation. NOCS\newcite{nocs} and CASS\newcite{cass}, for example, map different instances of each category into a unified representational space based on RGB or RGB-D features. However, the assumption of a unified space potentially leads to a decrease in performance in case of strong object variations. FS-Net\newcite{FS-Net} proposes an orientation-aware autoencoder with 3D graph convolutions for latent feature extraction where translation and scale are estimated using a tiny PointNet\newcite{pointnet}. Furthermore, Chen \etal\newcite{NeuralSynthesis} provide an alternative based on  ``analysis-by-synthesis" to train a pose-aware image generator, implicitly representing the appearance, shape and pose of the entire object categories. However, these methods require a pretrained object detector on each specific category which limits their generalization ability across categories.

In this paper, we present a new meta-learning based approach to increase the generalization capability of 6D pose estimation. To our knowledge, this is the first work that allows generalization across object categories. The main idea of our method lies in meta-learning object-centric representations in a category-agnostic way. Meta-learning aims to adapt rapidly to new tasks based only on a few examples. More specifically, we employ Conditional Neural Processes (CNPs)\newcite{CNP} to learn a latent representation of objects, capturing the generic appearance and geometry. Inference on new objects then merely needs a few labeled examples as input to extract a respective representation. In particular, fine-tuning on new objects is not necessary. A comparison between traditional instance-level approaches and ours is illustrated in \cref{fig:intro}. 

For feature extraction, we use FFB6D\newcite{ffb6d}, which learns representative features through a fusion network based on RGB-D images.  However, instead of directly using the extracted features for downstream applications, \ie segmentation and keypoint offsets prediction, we add CNP on top of the fusion network to further meta-learn a latent representation for each object. CNP takes in the representative features from a set of context images of an object, together with their ground-truth labels, and yields a latent representation. The subsequent predictions for new target images are conditioned on this latent representation.

\begin{figure}[t]
	\centering
		\includegraphics[width=\columnwidth]{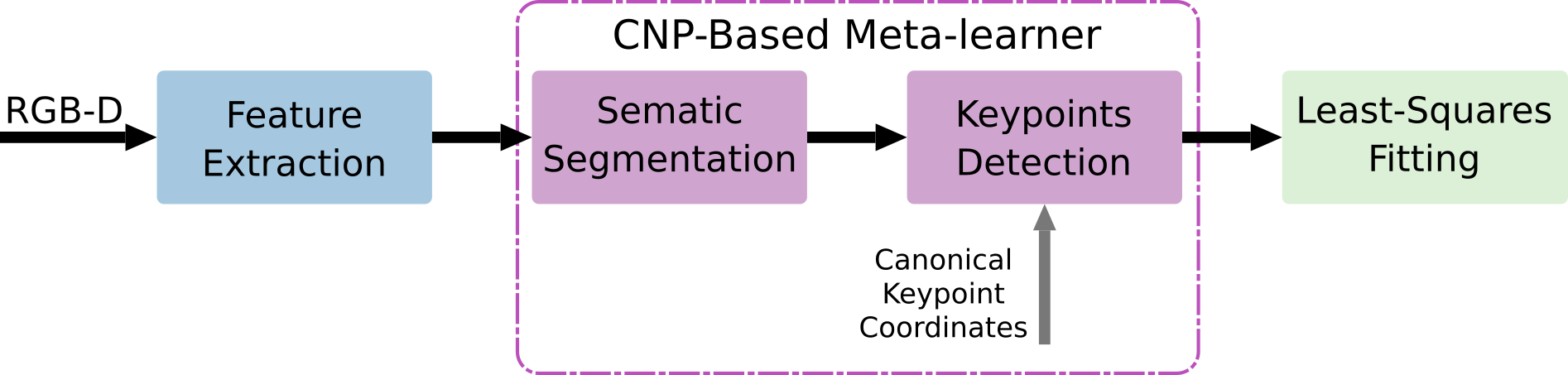}
			\caption[]{Schematic pipeline of our approach.}
			\label{fig:pipeline} 
	\end{figure}

To further leverage the object geometry and improve the keypoint prediction, we propose a novel GNN-based decoder which takes predefined canonical keypoints in the object's reference frame as an additional input and encodes local spatial constraints via message passing among the keypoints. Note that the additional input to the GNN does not require any further annotations on top of those existing datasets used by prior keypoint-based methods. The proposed pipeline is illustrated in \cref{fig:pipeline}.

Due to the lack of available data for cross-category level 6D pose estimation, we generate our own synthetic dataset for \textbf{M}ultiple \textbf{C}ategories in \textbf{M}ultiple \textbf{S}cenes (\textbf{MCMS}) using objects from ShapeNet\newcite{Chang2015ShapeNet} and extending the open-source rendering pipeline\newcite{blenderproc} with online occlusion and truncation checks. This provides us with the flexibility to generate datasets with limited and considerable occlusion respectively.
%Furthermore, we provide photorealistic scenes with limited and considerable occlusion respectively.

In summary, the main contributions of this work are as follows:
\begin{itemize}
    \item We introduce a novel meta-learning framework for 6D pose estimation with strong generalization ability on unseen objects within and across object categories.
    \item We propose a GNN-based keypoint prediction module that leverages geometric information from canonical keypoint coordinates and captures local spatial constraints among keypoints via message passing.
    \item We provide fully-annotated synthetic datasets with abundant diversity, which facilitate future research on intra- and cross-category level 6D pose estimation.
\end{itemize}

\section{Related Work}

 %---------------------------------------------
\noindent\textbf{6D Pose Estimation.} For instance-level 6D pose estimation, methods can be categorized into three classes: correspondence-based, template-based and voting-based methods\newcite{6d-review}.
Correspondence-based methods aim to find 2D-3D correspondences\newcite{dpod_2019_ICCV,hybridpose,pham2020lcd} or 3D-3D correspondences\newcite{StickyPillars_2021_CVPR}. Template-based methods, on the other hand, match the inputs to templates, which can be either explicit pose-aware images\newcite{texture-less2012,linemod-dataset} or templates learned implicitly by neural networks\newcite{Sundermeyer_2018_ECCV}. Voting-based approaches\newcite{pvnet,pvn3d,ffb6d} generate voting candidates from feature representations, after which the RANSAC algorithm\newcite{Fischler1981RandomSC} or a clustering mechanism such as MeanShift\newcite{meanshift} is applied for selecting the best candidates.
Our feature extractor, FFB6D\newcite{ffb6d}, falls into this latter category. FFB6D proposes a bidirectional fusion module to combine appearance and geometry information for feature learning. The extracted features are then used to predict per-point semantic labels and keypoint offsets, after which MeanShift is used to vote for 3D keypoints. Finally, the keypoints are used to predict the final 6D pose by Least-Squares Fitting\newcite{lsf}.

Recently, category-level 6D object pose estimation has gained increasing attention\newcite{nocs,cass,6-pack,FS-Net,NeuralSynthesis}. Wang \etal\newcite{nocs} share a canonical representation for all possible object instances within a category using Normalized Object Coordinate Space (NOCS). However, inferring the object pose by predicting only the NOCS representation is not easy given large intra-category variations\newcite{poseoverview}. To tackle this problem, \newcite{shapeprior} accounts for intra-category shape variations by explicitly modeling the deformation from shape prior to object model while CASS\newcite{cass} generates 3D point clouds in the canonical space using a variational autoencoder (VAE). FS-Net\newcite{FS-Net} proposes a shape-based model using 3D graph convolutions and a decoupled rotation mechanism to further reduce the sensitivity of RGB features to the color variations. However, these methods model the feature space explicitly on a category-level and therefore have a limited generalization ability across categories. By contrast, our method learns 6D pose estimation in a category-agnostic manner and can handle new objects from unseen categories.\\

\noindent\textbf{Meta-Learning.}
Meta-learning, also known as learning to learn, aims to acquire meta knowledge that can help the model to quickly adapt to new tasks with very few samples. In general, meta-learning can be categorized into metric-based\newcite{matching-network,prototypical,relation-network}, optimization-based\newcite{MAML,reptile,online-maml} and model-based\newcite{mann,CNP,NP,ANP} methods. Many meta-learning approaches have been applied to computer vision applications, \eg few-shot image classification\newcite{few-classification_2018,few-classification_2020_CVPR,few-classification-2,few-classification-3-ICLR2021}, vision regression\newcite{Gao_2022_CVPR}, object detection\newcite{few-obj_2020_CVPR,few-obj-2_2020_CVPR,few-obj-3_2021_CVPR,few-obj-4_2021_CVPR,few-obj-5_journal}, robotic grasping\newcite{gao2023metalearning},
semantic segmentation\newcite{few-seg-1_2019_ICCV,few-seg-2_2020_CVPR,few-seg-3-jounral-2021,few-seg-4-medical-2021} and 3D reconstruction\newcite{few-reconstruction-1_2019_ICCV,few-reconstruction-2_2020}. Our work is based on Neural Processes (NPs)\newcite{NP,ANP,fnp,CCNP,rnp_2020}, which fall into the category of model-based meta-learning approaches. NPs have shown promising performance on simple tasks like function regression and image completion. However, their application to 6D pose estimation has not yet been explored properly. We introduce CNP\newcite{CNP} to this problem in order to tackle the issue of poor generalization ability of existing methods on both intra- and cross-category level.\\

\noindent\textbf{Graph Neural Networks.} Graph neural networks (GNNs) have been widely applied on vision applications, such as image classification\newcite{classification-3,classification-2,classification-1}, semantic segmentation\newcite{semantic-4, semantic-gnn-1,semantic-gnn-2,semantic-gnn-3}, 
and object detection\newcite{det-1, det-2,det-3}. Recently, many works start using GNNs on human pose estimation\newcite{Graph-PCNN,Structure-aware-gnn,LearningDynamics-gnn-hand}.
Yang \etal\newcite{LearningDynamics-gnn-hand} derive the pose dynamics from historical pose tracklets through a GNN which accounts for both spatio-temporal and visual information while PGCN\newcite{Structure-aware-gnn} builds a directed graph over the keypoints of the human body to explicitly model their correlations. DEKR\newcite{DEKR} adopts a pixel-wise spatial transformer to concentrate on information from pixels in the keypoint regions and dedicated adaptive convolutions to further disentangle the representation.
Our approach is based on a similar idea as PGCN, where we take the keypoints in the canonical object coordinates as an additional input in order to leverage the spatial constraints between keypoints. We show that this drastically increases the performance on unseen objects and robustness on occluded scenes.

%%%%%%%%%%%%%%%%% 3. %%%%%%%%%%%%%%%%%%%%%
\section{Preliminary - Conditional Neural Processes}
Conditional Neural Processes (CNPs)\newcite{CNP} can be interpreted as conditional models that perform inference for some target inputs $x_t$ conditioned on observations, called ``contexts". These contexts consist of inputs $x_c$ and corresponding labels $y_c$ originating from one specific task. Note that in our case, each distinct object is considered as a task.

The basic form of CNP comprises three core components: 
encoder, aggregator and decoder.
The encoder takes a set of $M_c$ context pairs from a given task $C=\{(x_c^i, y_c^i)\}_{i=1}^{M_c}$ and extracts embeddings from each context pair respectively, $r_i = h_{\theta}(x_c^i, y_c^i), \ \forall (x_c^i,y_c^i) \in C$, where $h$ is a neural network parameterized by $\theta$. 
Afterwards, the aggregator $a$ summarizes these embeddings using a permutation invariant operator $\otimes$ %\eg max operator in our work, 
and yields the global latent variable as task representation:  $z = a(r_1,r_2,...,r_{M_c}) = r_1 \otimes r_2 \otimes... \otimes r_{M_c}$. 
Since the size of context set $M_c$ varies and the task representation has to be independent of the order of contexts, a permutation invariant mechanism is essential. Max aggregation is used in our model as we empirically find it outperforms mean aggregation, which is used in the original CNP. Finally, the decoder performs predictions for a set of target inputs $T=\{x_t^i\}_{i=1}^{M_t}$ conditioned on the corresponding task representation $z$ extracted and aggregated before: $\hat{y}_t^i = g_{\phi}(x_t^i,z), \ \forall x_t^i \in T$. $M_t$ is the number of target inputs, $g$ denotes the decoder, a neural network parametrized by $\phi$.

%For making $m$ predictions with $n$ observations, CNP only require a forward pass in the neural network, which scales with $\mathcal{O}(n+m)$, as opposed to $\mathcal{O}((n+m)^{3})$ of GPs.
% \comment{The strength of CNPs lies in their ability to extract representations for new tasks from very few context samples and make good predictions for corresponding target inputs.
% For 6D pose estimation, it is non-trivial to employ prior keypoint-based methods on new objects since each object includes different predefined keypoints,which are randomly sampled using \eg Farthest Point Sampling (FPS). Therefore, it is essential to capture keypoint representations from object features, \eg shapes and appearances using context.
% % Therefore they are well-suited for our goal of intra- and cross-category generalization.
% }
The ability to extract meaningful latent representation from very few samples renders CNP well-suited for our purposes. Due to the fact that each distinct object comes with different predefined keypoints, prior keypoint-based methods for 6D pose estimation do not generalize well to novel objects. Meta-training CNP to extract latent keypoint representations from object features, however, allows us to overcome this difficulty.
% In the simplest form of CNP, the encoder maps the labeled input pairs into some embeddings, which will be aggregated using a commutative operation. Through the decoder, the model yields the final prediction. Formally,

%%%%%%%%%% 4. %%%%%%%%%%%%%%%
\section{Approach}
In this paper, we propose a keypoint-based meta-learning approach for 6D pose estimation on unseen objects. Given an RGB-D image, the goal of 6D pose estimation is to calculate the rigid transformation $[R;t]$ from the object coordinates to the camera coordinates, where $R\in SO(3)$ represents the rotation matrix and $t\in\mathbb{R}^{3}$ represents the translation vector. We build on keypoint-based methods, that first predict the location of keypoints in camera coordinates from input RGB-D images and then regress the transformation between these and predefined keypoints in the object coordinates. The predefined keypoints in canonical object coordinates are thereby fixed beforehand, \eg using the Farthest Point Sampling (FPS) algorithm on the object mesh.

    \begin{figure*}[]%
    \centering
    \includegraphics[width=0.95\textwidth]{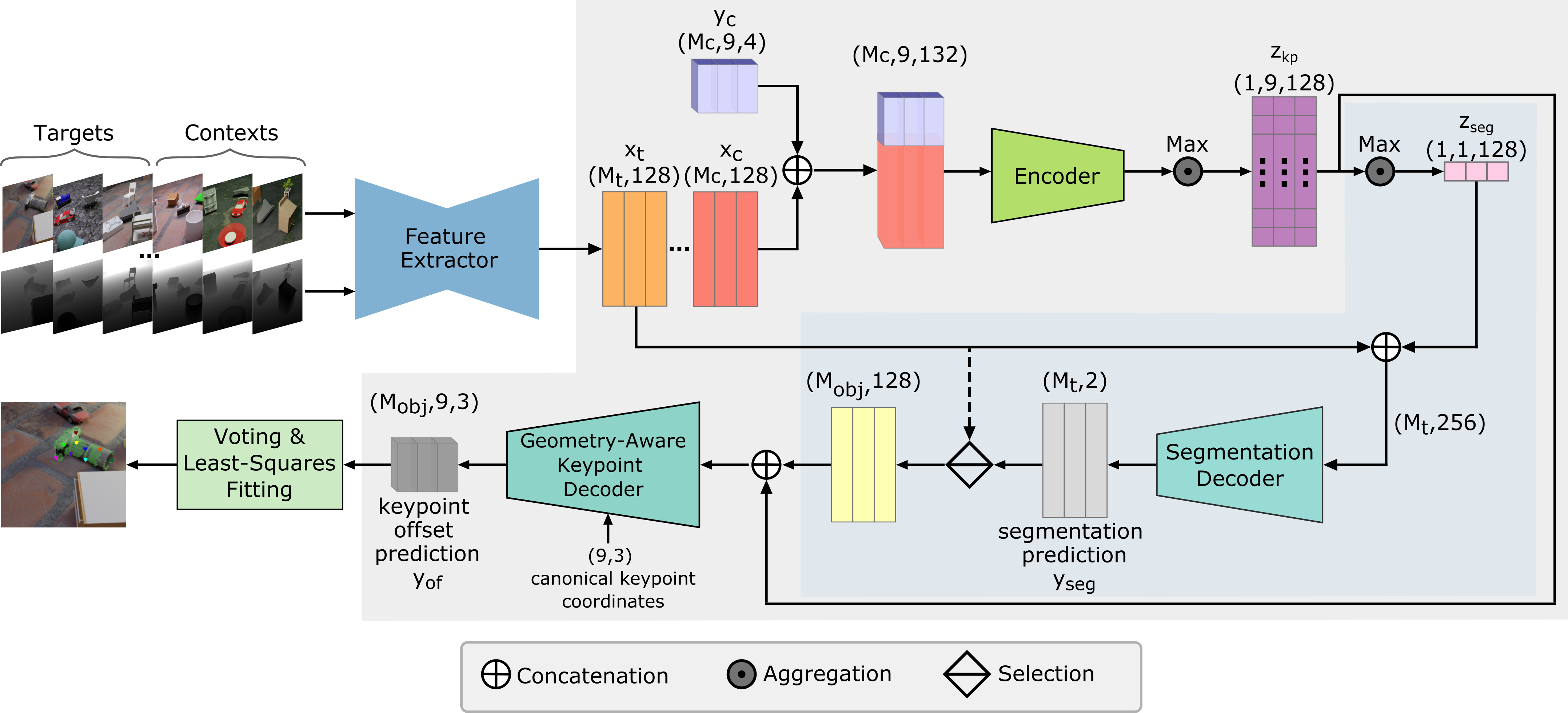}
    \caption[]{\textbf{Overview of the three stages of our method.} a) The feature extractor takes RGB-D images as inputs and produces point-wise features for a set of $M_c$ ($M_t$) points subsampled from the input context (target) image. b) The meta-learner (grey shaded area) encodes and aggregates the features of several context images into two latent variables $z_{kp}$ and $z_{seg}$. The segmentation module (blue shaded area) predicts a binary semantic label for each of the $M_t$ feature points of a target image conditioned on the latent representations $z_{seg}$, indicating whether the respective point belongs to the queried object. The keypoint decoder predicts per-point offsets for each keypoint based on the segmented features and the keypoint latent variables $z_{kp}$. 
    c) Lastly, 6D pose parameters are computed via voting and least-squares fitting.}
    \label{fig:overview}
    \end{figure*}

%%%%%%%%%% Overview %%%%%%%%%%
\subsection{Overview}
We consider 6D pose estimation in three stages: feature extraction, keypoint detection and pose fitting. At the first stage, we employ the feature extractor FFB6D\newcite{ffb6d} to extract representative features from RGB-D images.
% \comment{, which extracts appearance and geometry information from RGB-D images in order to learn representative features. }
For the second stage we use a CNP-based meta-learning approach.
%In our setting, representative features serve as inputs $x$, per-point translation offsets w.r.t keypoints and segmentation labels as outputs $y$.  
%As can be seen in  \cref{fig:overview}, the input of RGB-D images is split into a context and a target set.
The flow of context and target samples through our model is shown in  \cref{fig:overview}, where the context inputs for each task $x_c$, \ie the features extracted from the context RGB-D images, and the correpsonding labels $y_c$ are used jointly to distill a task representation. This representation serves as prior knowledge for the subsequent prediction on target inputs $x_t$.
%More specifically, extracted features of context images  along with their corresponding ground-truth keypoint offsets and segmentation labels are considered as observations for CNP. 
We use two decoders in our meta-learning framework, predicting semantic labels and 3D keypoint offsets respectively. Furthermore, we propose a novel geometry-aware decoder using a GNN for the keypoint offsets prediction, which explicitly models the spatial constraints between the keypoints. Finally, the 6D pose parameters are regressed by least-squares fitting at the third stage. %We will now describe the core components of our method in more detail.

\subsection{Feature Extraction}
For feature extraction we rely on the fusion network FFB6D\newcite{ffb6d} which combines appearance and geometry information from RGB-D images and extracts representative features for a subset of seed points sampled from the input depth images. Therefore, the output is a set of per-point features corresponding to the sampled seed points.

\comment{This model takes RGB-D images as input and extracts representative features for a subset of seed points sampled from the input depth image. Its output is a tensor of per-point features. }

%%%%%%%%%% NP Main Part %%%%%%%%%%
\subsection{Meta-Learner for Keypoint Detection}
% \comment{To improve the generalization performance, we introduce CNP-based meta-learner for estimating keypoints in a category-agnostic manner. }
Two steps are involved in the keypoint estimation procedure: segmentation of the queried object and keypoint detection, which both rely on a preceding extraction of latent representations.
%All these operations will be explained in the following. 
% \comment{Two steps are involved in the detection procedure: Firstly, the meta-learner identifies and segments the queried object using a conditional module. Secondly, point-wise translation offsets w.r.t. each keypoint are predicted based on the selected seed points on the surface of queried object, following by a clustering algorithm in order to determine the 3D keypoint coordinates.
% }
% The learner captures the target feature and geometry from context instead of memorizing fixed number of known objects, that's where meta-learning comes in and differs with existing methods.

%%%%%% Segmentation %%%%%%
\textbf{Extraction\comment{and aggregation} of latent representations.}
Identifying and distinguishing a novel object from a multi-object scene and extracting its keypoints requires modules, which are  conditioned on the latent representation of the queried object. In order to obtain such a latent representation, we need a\comment{well-aligned} set of context samples $\{(x_{c, i}, y_{c, i})\}_{i=1}^{M_c}$. Here $x_{c, i}$ denotes the per-point features extracted in the first stage from context images and $y_{c, i}=\{y_{c,i}^u\}_{u=1}^{M_K}$ is the ground-truth label where $y_{c, i}^u=\{y_{of}^u, y_{seg}\}_{c, i}$ includes the 3D keypoint offsets $y_{of}^u$ between the seed point and predefined keypoint $p_u$, and semantic label $y_{seg}\in \{0, 1\}$ indicating whether the seed point belongs to the queried object. Given a context sample as input, an encoder generates per-seed-point embeddings 
for each of the $M_k$ keypoints to be predicted:
%for each  keypoint prediction:
\begin{align} \label{eq:encoder}
r_{i}^{u} = h_{\theta}(x_{c, i} \oplus y_{c, i}^u),\  i=1,...,M_c,\ u=1,..., M_k,
\end{align}
where 
%$x_c$ represents the output of the fusion network, containing the extracted features of a subset of $M_c$ seed points of the input depth image.
$M_c$ denotes the number of seed points selected from each context image; $M_k$ is the number of selected keypoints which in our case is 9.
$\oplus$ stands for the concatenation operation, where the inputs are first broadcast to the same shape, if necessary.
%Note that when two inputs have different dimensions, the one with the lower dimension will be repeated and broadcast to the same dimension as the higher one.
The obtained embeddings are next aggregated by max aggregation to first obtain a latent representation $z_{kp}^u$ for each keypoint. A second aggregation over these keypoint representations is then applied in order to extract a representation $z_{seg}$ for the segmentation task: %We acquire latent variables for offsets and segmentation prediction respectively by applying max aggregation on two different levels over embeddings
% \begin{align}
%     z_{kp}^u &= \max\limits_{i=1}^{M_c}( r_{i}^{u} ), \ u=1,..., M_k \label{eq:z-kp} \\
%     z_{seg} &= \max\limits_{u=1}^{M_{k}}( z_{kp}^u)\label{eq:z-seg}
% \end{align}\label{eq:z}

\begin{equation}
\label{eq:z-kp}
    z_{kp}^u = \max\limits_{i=1}^{M_c}( r_{i}^{u} ), \ u=1,..., M_k, 
\end{equation}
\begin{equation}
\label{eq:z-seg}
    z_{seg} = \max\limits_{u=1}^{M_{k}}( z_{kp}^u).
\end{equation}
%where $z_{kp}^u$ denotes the global latent variable for the $u$-th keypoint prediction, and $z_{seg}$ for the segmentation.

\textbf{Conditional Segmentation.} In the step described above, the model encapsulates relevant information (\eg shape and texture attributes) into the latent variable $z_{seg}$. This can then be used to identify and locate the queried object in the target images. 
The segmentation decoder  $g_{\mathcal{S}}$ takes the latent variable $z_{seg}$ and features $x_t$ extracted from the target images (see  \cref{fig:overview}) and predicts a semantic label for each seed point via a multi-layer perceptron (MLP):  
\begin{align} \label{eq:decoder-seg}
y_{seg, i}= g_{\mathcal{S}}(x_{t, i} \oplus z_{seg}), \ i=1,...,M_t,
\end{align}
where $M_t$ is the number of seed points sampled from each target image, $x_{t, i}$ denotes the corresponding extracted features. These per-point segmentation predictions $y_{seg}$
% the features of the target image is filtered by only selecting the seed points which
are then used to select only the seed point features $x_{obj}$ belonging to the queried object from $x_{t}$
for the subsequent keypoint prediction.

%%%%%% Keypoints Detection %%%%%% 
\textbf{Conditional Keypoint Offset Prediction.} 
%Similar to segmentation, the per-point translation offsets are meta-learned given the embedded keypoint information. More specifically, 
The keypoint offsets decoder  $g_{\mathcal{K}}$ takes the features extracted by the segmentation module along with the latent variables $z_{kp}$ as input and predicts translation offsets $y_{of}$ for each keypoint:
\begin{align} \label{eq:decoder-kp}
y_{of, i}^u &= g_{\mathcal{K}}(x_{obj, i} \oplus z_{kp}^{u}),\notag\\
&i=1,...,M_{obj},\  u=1,..., M_k,
\end{align}
where $M_{obj}$ denotes the number of selected seed points on the queried object, $x_{obj, i}$ denotes the object features of $i$-th seed point. 
%$M_{obj}$ denotes the number of selected seed points.
The decoder $g_{\mathcal{K}}$ can be any appropriate module in \cref{eq:decoder-kp}. In the vanilla version of our framework, it is given by a trivial MLP. However, we use a GNN for $g_{\mathcal{K}}$ in our final version, the details of which will be given in \cref{methods:gnn}.

\textbf{Pose Fitting.} 
%%%%%%%%%% Least-Squares %%%%%%%%%%
% \subsection{Keypoint-Based 6D Pose Estimation}
Similar as in \newcite{ffb6d}, we adopt MeanShift\newcite{meanshift} to obtain the final keypoint  prediction $\{p_i^{*}\}_{i=1}^{M_k}$ in the camera coordinates, based on  keypoint candidates output by the keypoint decoder.
Given predefined 3D keypoints in object coordinates $\{p_i\}_{i=1}^{M_k}$, 6D pose estimation can be converted into a least-squares fitting problem\newcite{lsf} where the optimized pose parameters $[R;t]$ are calculated by minimizing the squared loss using singular value decomposition(SVD):
    \begin{align} \label{eq:lsf-fitting}
    L_{lsf} = \sum_{i=1}^{M_k} \norm{p_i^{*} - (R\cdot p_i + t)}^2 .
    \end{align}

%%%%%% GNN %%%%%% 
\subsection{Geometry-Aware Keypoint Decoder}
\label{methods:gnn}
\begin{figure}[t]
\centering
\includegraphics[width=0.77\columnwidth]{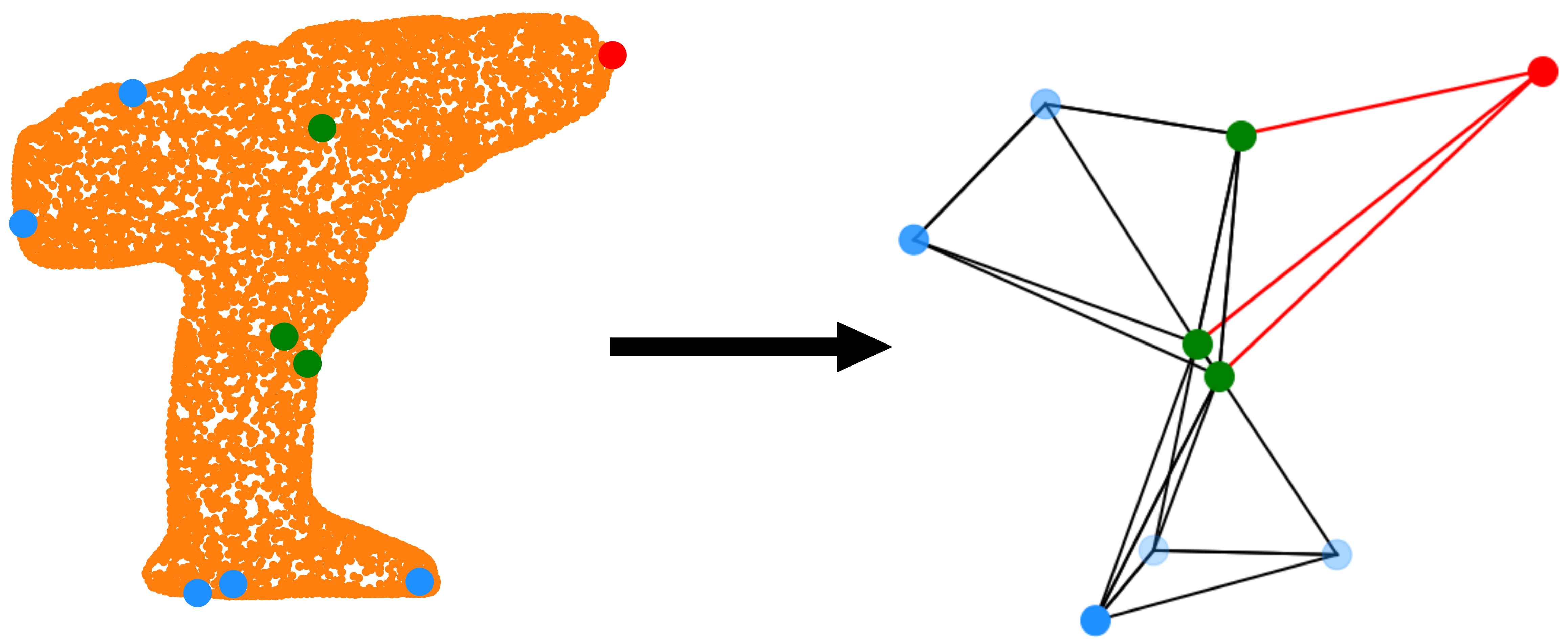}
%{latex/figs/gnn_driller.png}
	\caption[]{\textbf{Example of the graph generation.} The node positions are determined by the predefined keypoints in object coordinates. By applying the K Nearest Neighbor (KNN) algorithm, we find the k closest adjacent nodes of each parent node and connect them by edges. For instance, the given graph is generated with $k=3$, where the red node is selected as the parent node and green nodes are the three nearest neighbors. The driller is sampled from LineMOD.}
	\label{fig:gnn_driller} 
\end{figure}

%\textbf{Keypoint-Based Spatial Attention}
Similar to prior methods\newcite{pvn3d,ffb6d}, we rely on predefined object keypoints for the final pose fitting. %In addition 
However, we also utilize them as an additional input to the keypoint decoder. Since they contain useful prior knowledge of the object's geometric structure, they can significantly improve keypoint detection. 
In order to highlight the additional input to our decoder, we rewrite \cref{eq:decoder-kp} as follows:
% \begin{align} \label{eq:decoder-kp-gnn}
% y_{of}^{i} &= g_{\mathcal{K}}(x_{obj}, \bigcup_{j\in\mathcal{N}(i)}z_{kp}^{i}, \bigcup_{j\in\mathcal{N}(i)}p_j)
% \end{align}
\begin{align} \label{eq:decoder-kp-gnn}
y_{of,i}^{u} &= g_{\mathcal{K}}(x_{obj,i},\, z_{kp}^{v},\, p_v), \  v\in \mathcal{N}(u),
\end{align}
where $\mathcal{N}(u)$ denotes the neighbor set of keypoint $u$ including $u$ itself and $p_v$ are the 3D object coordinates of keypoint $v$. To leverage the geometric information contained in the relation among the keypoints, we propose a GNN-based decoder $g_{\mathcal{K}}$ instead of the trivial MLP in \cref{eq:decoder-kp}. For this purpose, we create a graph over the keypoints of each object. The nodes are given by the keypoints which share edges with their $k$ nearest neighbours. \cref{fig:gnn_driller} illustrates an example with $k=3$.
Internally, \cref{eq:decoder-kp-gnn} is split into the following two steps involved in message passing along the graph:
% \begin{align}
% \alpha_{i}^{u, v} &= f^{l} \left(x_{obj,i} \oplus z_{kp}^{v}, p_u - p_v \right), \  \forall v\in \mathcal{N}(u) \label{eq:gnn-1}\\
% y_{of,i}^{u} &= f^{g} \left(
% \max_{v\in \mathcal{N}(u)} \ \alpha_{i}^{u, v} \right) \label{eq:gnn-2}
% \end{align}
\begin{equation}
\label{eq:gnn-1}
\alpha_{i}^{u, v} = f^{l} (x_{obj,i} \oplus z_{kp}^{v}, p_u - p_v ), \  \forall v\in \mathcal{N}(u),
\end{equation}
\begin{equation}
 \label{eq:gnn-2}
    y_{of,i}^{u} = f^{g} (\max_{v\in \mathcal{N}(u)} \ \alpha_{i}^{u, v} ).
\end{equation}
$g_{\mathcal{K}}$ is correspondingly composed of two sub-networks, $f^l$ and $f^{g}$. These correspond to updating the messages $\alpha_{i}^{u, v}$ sent along all edges, aggregating the messages arriving at each node $u$ to update the corresponding node features %$\alpha_i^u$
and decoding them into keypoint offsets $y_{of,i}^{u}$. 
%The local mapping $f^l$ maps the features of each node's neighbors to messages taking into account their spatial relation,  while the global network $f^{g}$ integrates all messages arriving at each node and predicts the final keypoint offsets.

\comment{
    \begin{align} \label{eq:gnn-decoder-offsets}
        x_i^{(q+1)} = \gamma_{\Theta} \left(
        \max_{j\in \mathcal{N}(i)\cup\{i\}} \ \psi_{\Theta} \left(x_j^{(q)}, p_j - p_i \right) \right),
    \end{align}
}

\section{Experiments}
%\resizebox{\textwidth}{!}{
% Second version of table, with booktabs.

%%%%%%%%%%%%%%%%%%%%%%%%%%%
\subsection{Datasets}
\begin{figure}[htb!]
  \centering
  \begin{subfigure}{0.31\columnwidth}
      \includegraphics[width=\columnwidth]{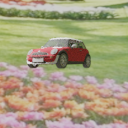}
    \caption{Toy}
    \label{fig:data-toy}
  \end{subfigure}
  %\hfill
  \begin{subfigure}{0.31\columnwidth}
      \includegraphics[width=\columnwidth]{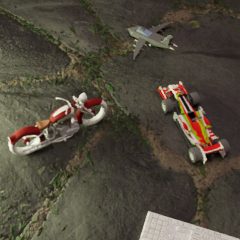}
    \caption{PBR}
    \label{fig:data-pbr}
  \end{subfigure}
  %\hfill
    \begin{subfigure}{0.31\columnwidth}
    \includegraphics[width=\columnwidth]{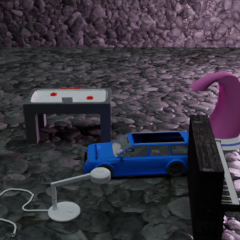}
    \caption{Occlusion}
    \label{fig:data-occ}
  \end{subfigure}
  \caption{Samples from MCMS dataset.}
  \label{fig:our-data}
\end{figure}

%%%% Large Figures! %%%%%
\begin{figure*}[]%
\centering
\includegraphics[width=0.95\textwidth]{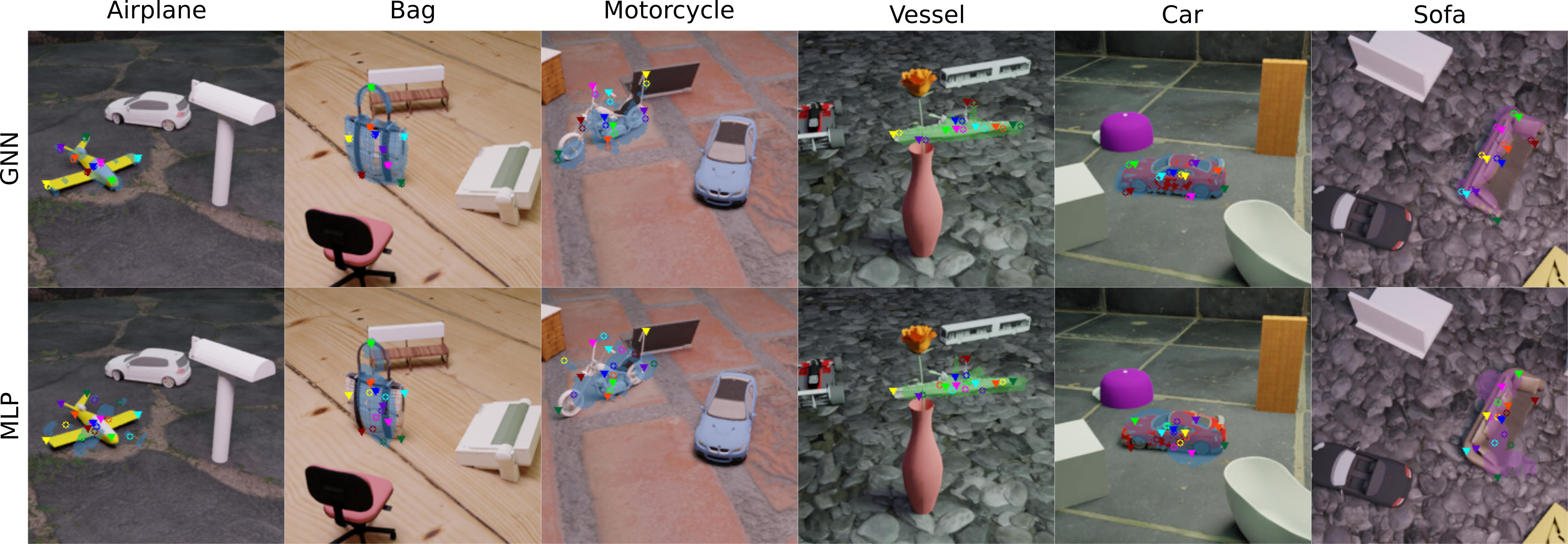}
\caption[]{\textbf{Qualitative comparison between GNN and MLP decoder for keypoints prediction on PBR-MCMS.} Triangles and circles are the projected ground-truth and predicted keypoints respectively. The keypoint predictions of the MLP decoder are randomly shifted without considering geometric constrains between keypoints. By contrast, the predictions by the GNN decoder are more accurate. The example of the motorcycle shows that though the keypoints predicted by the GNN are slightly shifted here, the geometric constraints are met, resulting in a uniform shift of all keypoints.}
\label{fig:gnn_vs_mlp}
\end{figure*}
%Object points in object coordinates are transformed by the predicted pose and project to the image plane using camera intrinsic parameters, which are visualized in the examples.

\textbf{LineMOD.} LineMOD\newcite{linemod-dataset} is a widely used dataset  for 6D pose estimation which comprises 13 different objects in 13 scenes. Each scene contains multiple objects, but only one of them is annotated with a 6D pose and instance mask.

%Multi-category Multi-scene (MCMS dataset) 
\textbf{MCMS dataset.} Due to the unavailability of datasets for cross-category level 6D pose estimation, we generate two fully-annotated synthetic datasets using objects from ShapeNet\newcite{Chang2015ShapeNet}, which contain various objects from \textbf{Multiple Categories} in \textbf{Multiple Scenes} \textbf{(MCMS)}.
%where the photorealistic version can be further divided into  non-occluded and occluded one.
%thus we name our datasets as namely Toy-, PBR-, and Occlusion-MCMS. 
The simple version of MCMS, named Toy-MCMS, is composed of images containing a single object with backgrounds randomly sampled from the real-world image dataset SUN \newcite{suncg}. Our second dataset can be further divided into a non-occluded and an occluded version, called PBR-MCMS and Occlusion-MCMS. To create these datasets, we extend the open-source physics-based rendering (PBR) pipeline\newcite{blenderproc} with functionalities such as online truncation and occlusion checks. For each image, five objects are placed in a random scene with textured planes and varying lighting conditions. Images are then photographed with a rotating camera from a range of distances. PBR-MCMS contains images without occlusion while Occlusion-MCMS contains images with $5\%-20\%$ occlusion of the queried object. \cref{fig:our-data} shows an example for each dataset using  an object from the car category as the queried object.% \tocheck{T.B.C: every image has occlusion }

%%%%%%%%%%%%%%%%%%%%%%%%%%%
\subsection{Evaluation Metrics}
We use the average distance metrics ADD\newcite{linemod-dataset} for evaluation. Given the predicted 6D pose $[R;t]$ and the ground-truth pose $[R^{*};t^{*}]$, the ADD metric is defined as:
\begin{align} \label{eq:add-metric}
\mathrm{ADD}=\frac{1}{m}\sum_{x \in \mathcal{O}}\norm{(Rx+t)- (R^{*}x+t^{*})},
\end{align}
where $\mathcal{O}$ denotes the object mesh and $m$ is the total number of vertices on the object mesh. This metric calculates the mean distance between the two point sets transformed by predicted pose and ground-truth pose respectively. Similar to other  works\newcite{posecnn,pvnet,ffb6d}, we report the ADD-0.1d accuracy, which indicates the ratio of test samples, where the ADD is less than 10\% of the object's diameter. 
% \comment{
% In general, the average distance ADD metric is used for non-symmetric objects whereas ADD-S for symmetric objects. Since the matching between points is ambiguous for some poses, ADD-S computes the mean distance based on the closest point distance:
% \begin{align} 
% \mathrm{ADD-S}=\frac{1}{m}\sum_{x_1 \in \mathcal{O}} \min_{x_2\in\mathcal{O}}\norm{(Rx+t)- (R^{*}x+t^{*})}
% \end{align}\label{eq:adds-metric}
% }

%%%%%%%%%%%%%%%%%%%%%%%%%%%
\subsection{Implementation and Training Details}
For each object, we define 9 keypoints, where 8 keypoints are sampled from the 3D object model using FPS, and the other one is the object center. The nearest neighbors used for each keypoint is set to $k=8$ in our geometry-aware decoder. To train the meta-learner, we use the Focal Loss\newcite{focal-loss} to supervise the segmentation module and a L1 loss for per-point translation offset prediction. The overall loss is weighted sum of both terms, with a weight 2.5 for segmentation and 1.0 for keypoint offsets. During training, for each iteration, we arbitrarily sample 18 objects and 12 images per object. The number of context images is randomly chosen between 2 and 8 per object while the remaining images are used as target set. 

\textbf{Training setup.} For the \linemod dataset, we use iron, lamp, and phone as novel objects for testing and the 10 remaining objects for training. Since \linemod contains only a very limited number of  objects, we only evaluate the keypoint offset prediction module using the ground-truth segmentation for selecting the points belonging to the queried object. For Toy- and PBR-MCMS, we use 20 and 19 categories for training respectively, with 30 objects per category and 50 images per object. During evaluation, 30 novel objects of each training category are tested for intra-categorical performance and 5 novel categories for cross-category performance. All experiments are conducted on NVIDIA V100-32GB GPU.
%More details regarding category split are shown in \cref{tab:full-TOY-multi} and \cref{tab:full-PBR-multi}. 

%%% Large figures! %%%

%%% Large Tables %%%%%%
\begin{table*}[hbt!]
%\centering
    \begin{minipage}{.38\linewidth}
      \centering
\begin{tabular}{@{}l|c|c@{}}\toprule
& \multicolumn{1}{c|}{Vanilla-ML} & \multicolumn{1}{c}{GAML}
\\\midrule
%\\\cmidrule(lr){2-3}\cmidrule(lr){4-5}
Category & ADD & ADD    
\\\midrule
Airplane   & 80.6 & \textbf{87.2} \\
Bench     & 56.7  & \textbf{72.3}\\
Chair     & 62.8  & \textbf{80.9}\\
Motorcycle  & 92.6 & \textbf{94.7} \\
Washer  & 85.4 & \textbf{91.4} \\
Bus*  & 83.0  & \textbf{85.4} \\
Cap*  & 46.6 &\textbf{54.2}  \\
Laptop*  & 18.8   &\textbf{48.8} \\
Piano*  & 47.1 & \textbf{50.7}   \\
Remote*  & 53.5 & \textbf{56.1} \\
\midrule
Intra-Categ. & 74.2 & \textbf{81.9}  \\
Cross-Categ.  & 50.3 & \textbf{59.0} \\
All  & 69.4 & \textbf{77.2} \\
\bottomrule
\end{tabular}    
    \renewcommand\thetable{1}
    \caption{Multi-category evaluation on Toy-MCMS dataset. Novel objects are marked with *.} \label{tab:TOY-multi}  
\end{minipage}%
    \quad 
\begin{minipage}{.6\linewidth}
      \centering
    \begin{tabular}{@{}l|c|ccc|ccc@{}}\toprule
 & \multicolumn{1}{c|} {FFB6D} & \multicolumn{3}{c|} {Vanilla-ML} & \multicolumn{3}{c}{GAML}
\\\midrule
%\\\cmidrule(lr){2-3}\cmidrule(lr){4-5}
Category  & PBR  & PBR & Occ. & $\Delta$ & PBR & Occ. & $\Delta$      
\\\midrule
Airplane  & 9.1 &\textbf{90.4} &  43.2 & 47.2 & 89.8 & \textbf{46.6} & \textbf{43.2}  \\
Bench & 2.9 & 62.1 & 40.4 & 21.7 & \textbf{69.8} & \textbf{49.0} &\textbf{20.8}  \\
Chair & 1.1 & \textbf{80.0} &54.4 & 25.6 & \textbf{80.0}& \textbf{55.6} &\textbf{22.4}   \\
Motorcycle & 12.7 &\textbf{90.2} & \textbf{64.8 }& \textbf{25.4}   & 85.6 & 54.4 & 31.2  \\
Washer & 4.1  &54.8 & 37.7 &  17.1 & \textbf{68.1}  & \textbf{55.0} &\textbf{13.1} \\
Birdhouse* & 0.8 & \textbf{35.6} & 23.5 & 12.1& 35.4 & \textbf{28.0} & \textbf{7.4} \\
Car* & 2.4 & 52.5 & 42.9 &  \textbf{9.6} & \textbf{56.9} & \textbf{44.4} & 12.5   \\
Laptop* & 1.3 &54.0 & 26.0 & \textbf{28.0} & \textbf{85.0} & \textbf{47.7} & 37.3  \\
Piano*& 2.0 & \textbf{45.8} &27.3 & 18.5 & \textbf{45.8}& \textbf{32.5} & \textbf{13.3}   \\
Sofa* & 2.6 & 68.1 &45.4 & 22.7 & \textbf{69.8}  & \textbf{57.9} & \textbf{11.9} \\
\midrule
Intra-Categ. & 4.53 &58.2 & 38.7 & 19.5 & \textbf{62.9} & \textbf{43.9} & \textbf{19.0}  \\
Cross-Categ.& 1.81 &51.2&33.0& 18.2 &\textbf{58.6} & \textbf{42.1} & \textbf{16.5}  \\
All  &  3.96 & 56.7 & 37.6 & 19.1 & \textbf{62.0}& \textbf{43.5} & \textbf{18.5}\\
\bottomrule
    \end{tabular}
    \renewcommand\thetable{2}
    \caption{Multi-category evaluation on PBR- and Occlusion-MCMS datasets.  $\Delta$ represents the performance gap between PBR- and Occlusion-MCMS.}
    \label{tab:PBR-multi}
    \end{minipage} 
\end{table*}%\caption{Global caption}

%%%%%%%%%%%%%%%%%%%%%%%%%%%
\begin{table}[htb!]
    \centering
\begin{tabular}{@{}l|cc|cc@{}}\toprule
& \multicolumn{2}{c|}{FFB6D} & \multicolumn{2}{c}{Ours}
\\\midrule
%\\\cmidrule(lr){2-3}\cmidrule(lr){4-5}
Object & L1 Loss  & ADD & L1 Loss  & ADD      
\\\midrule
Ape & 0.06 & \textbf{100}  & 0.02&\textbf{100} \\
Holepuncher    & 0.07 & \textbf{100}    & 0.02 & \textbf{100}  \\
Iron*               & 1.39 & 0.6  & 0.26 &\textbf{36.2}  \\
Lamp*               & 1.52 & 0.5 & 0.38 & \textbf{22.4} \\
Phone*             & 0.89 & 0.0  & 0.17 & \textbf{17.8}  \\
\bottomrule
\end{tabular}
\renewcommand\thetable{3}
    \caption{Evaluation results on LineMOD dataset. %Novel objects are marked with a *.
    }
    \label{tab:LineMod}
\end{table}
%%%%%%%%%%%%%%%%%%%%%%%%%%%%%%
\subsection{Evaluation Results}
We evaluate our approach using the LineMOD and MCMS datasets at intra- and cross-category levels. More quantitative and qualitative results are provided in \cref{sec:appendix-res}.

%%%%%%%%%%%%%%%%
% \begin{table*}[]
%     \centering
% \begin{tabular}{lcccccl}\toprule
% & \multicolumn{2}{c}{Trained Classes} & \multicolumn{3}{c}{New Classes}
% \\\cmidrule(lr){2-3}\cmidrule(lr){4-6}
%           & Ape  & Holepuncher & Iron  & Lamp & Phone     \\\midrule
% FFB6D & 0.063/100 & 0.065/100 & 1.53/0 & 1.55/0 & 0.89/0 \\
% Ours  & 0.019/100 & 0.021/100 & 0.26/36 & 0.38/22 & 0.17/17  \\\bottomrule
% \end{tabular}
%     \caption{Evaluation results on LineMOD dataset. (L1 loss[m]/ ADD-0.1d[\%] are reported where * denotes novel objects.)}
%     \label{tab:LineMod}
% \end{table*}

\textbf{LineMOD.}  \cref{tab:LineMod} shows training and test results following \newcite{ffb6d}. Note that the segmentation ground-truth is used for these results and we only evaluate the performance and generalization ability of the keypoint offset prediction module. Our model not only performs better on training objects, but also generalizes well to new objects even though it is trained on a limited number of objects and tested on new objects with large variations in appearance and geometry.

%%%%%%%%%%%%%%%%
\textbf{Toy- \& PBR-MCMS.} \cref{tab:TOY-multi} shows test results on the Toy-MCMS  dataset, which demonstrate that our proposed GNN decoder (GAML) consistently outperforms the Vanilla Meta Learner (Vanilla-ML) using classic MLP decoder on all categories. \cref{fig:gnn_vs_mlp} visualizes some test examples for qualitative comparison. Next, we compare our meta-learner to FFB6D on the PBR dataset.  \cref{tab:PBR-multi} shows that our model generalizes well while FFB6D cannot directly transfer to novel objects. For a fair comparison, we further train FFB6D on the PBR dataset and fine-tune the pretrained model on each specific novel object with the same context images as given to GAML. \cref{tab:fine-tune} shows that our model still outperforms the fine-tuned FFB6D reliably and requires no trade-off between new and preceding tasks, whereas fine-tuning normally leads to a performance decrease on the previous tasks.

%%%%%%%%%%%%%%%%
% \textbf{Photorealistic dataset --- multi-category training}

\begin{table}[t]
  \centering
  \begin{tabular}{@{}c|cccccc@{}}
    \toprule
 &Airplane &Chair &Car  &Laptop &Sofa  \\
    \midrule
    FFB6D & 60.0 & 52.0 & 36.7 & 48.0 & 49.3 \\
    GAML & \textbf{89.8} & \textbf{80.0} & \textbf{56.9} & \textbf{85.0} & \textbf{69.8}  \\    
    \bottomrule
  \end{tabular}
  \caption{Comparison between GAML and fine-tuned FFB6D on PBR-MCMS using ADD metric.}
  \label{tab:fine-tune}
\end{table}
% Motorcycle 76.0  85.6

\begin{figure*}[]%
\centering
\includegraphics[width=\textwidth]{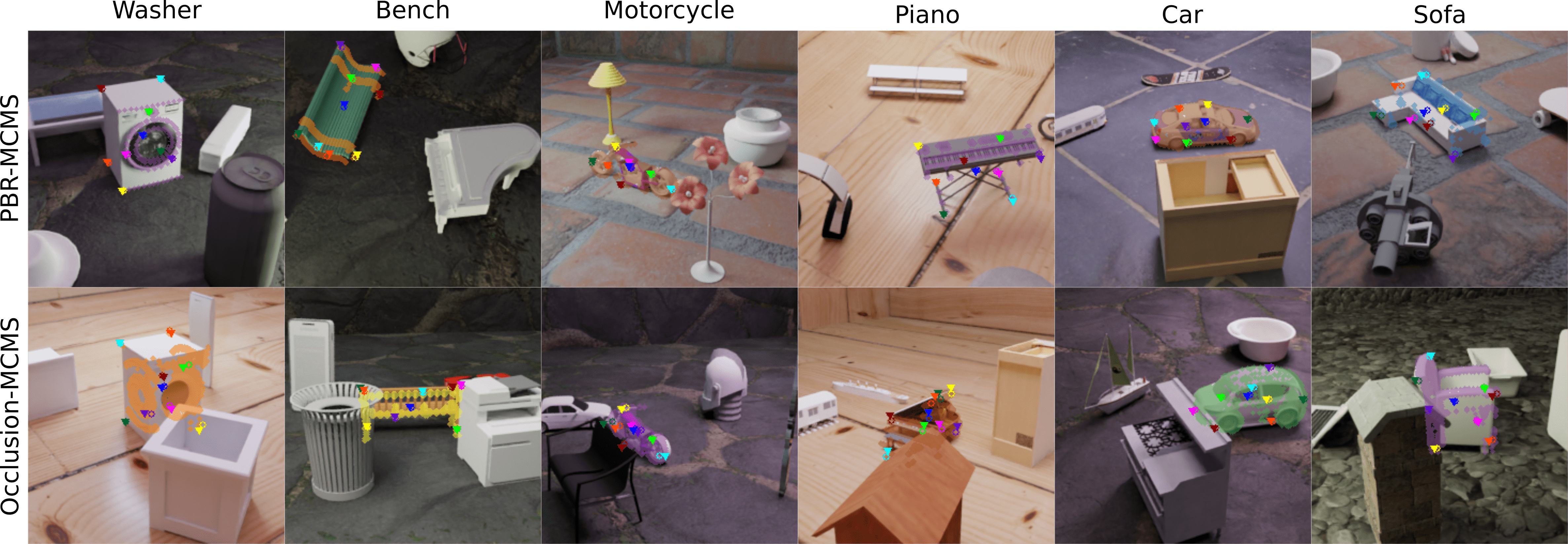}
\caption[]{\textbf{Qualitative results on PBR- and Occlusion-MCMS datasets.} Triangles and circles are the projections of ground-truth and predicted keypoints respectively. Note that our model is trained only on PBR-MCMS but shows robust performance on Occlusion-MCMS.}
\label{fig:PBR-occlusion}
\end{figure*}

%%%%%%%%%%%%%%%%
\textbf{Occlusion-MCMS.}   
Quantitative and qualitative results on Occlusion-MCMS  are presented in 
\cref{tab:PBR-multi} and \cref{fig:PBR-occlusion}. Strikingly, our approach achieves consistent and robust performance on occluded scenes even though training is conducted on non-occluded PBR-MCMS.

\comment{
\begin{table}[htb!]
  \centering
  \begin{tabular}{@{}c|c|c|c@{}}
    \toprule
       & ADD-0.1d &  & ADD-0.1d\\
    \midrule
    Airplane & 46.6 & Birdhouse* & 28.0 \\
    Bench &  49.0  & Car* & 40.6 \\
    Chair &  55.6  & Laptop* & 45.4\\
    Motorcycle & 54.4 & Piano* & 32.3\\
    Motorcycle &  46.5  & Sofa* & 51.7\\
    \bottomrule
  \end{tabular}
  \caption{Photorealistic occlusion  dataset}
  \label{tab:PBR-occlusion}
\end{table}
}

%%%%%%%%%%%%%%%%%%%%%%%%%%%
\subsection{Ablation Study}
\textbf{Effect of K Neighbors in GNN.} In \cref{tab:ablation-knn}, we study the effect of the $k$ neighbors in the GNN. We run tests using five seeds and calculate the mean. Compared to $k=3$, using all keypoints as neighbors can improve the robustness. We find this to be more crucial when training on a single category with limited object variations, where involving all keypoints gives more expressive spatial representation. Full statistics are presented in \cref{subsec:appendix-ablation-k}.

\begin{table}[htb!]
  \centering
  \begin{tabular}{@{}c|cc@{}}
    \toprule
       & k = 3 & k = 8\\
    \midrule
    %Intra-Categ. & 44.9 & 62.9  \\
    %Inter-Categ. &  42.4 & 58.6 \\
    %All &  44.4  & 62.0 \\
    Multi-Categ. & 60.1 & \textbf{61.9}  \\
    Single-Categ. &  77.9 & \textbf{83.3} \\
    \bottomrule
  \end{tabular}
  \caption{ADD Results  on PBR-MCMS using different number $k$ of neighbors in GNN decoder.}
  \label{tab:ablation-knn}
\end{table}

\textbf{Effect of the Aggregation Module in CNP.} In our work, CNP uses max aggregation instead of mean as used in the original paper\newcite{CNP}. We further compare max aggregation with the cross-attention module proposed in Attentive Neural Processes (ANPs)\newcite{ANP} removing the self-attention part. The training curves (see \cref{subsec:appendix-ablation-attention}) show that both methods achieve similar training performance, though ANP converges faster at the beginning. Nevertheless,  \cref{tab:ablation-aggregation} illustrates that CNP generalizes slightly better to novel tasks on both intra- and cross-category levels.

\textbf{Robustness to Occlusion.} To further illustrate the benefits coming from the geometry-aware estimator, we compare GAML with Vanilla-ML. The results in \cref{tab:PBR-multi} show that our purposed GNN decoder significantly improves the performance and robustness on occluded scenes.

\begin{table}[]
  \centering
  \begin{tabular}{@{}c|ccc@{}}
    \toprule
       & Intra-Categ. & Cross-Categ. &All\\
    \midrule
    %MEAN & 80.6 & 61.0  &  76.5 \\
    CNP & \textbf{81.9} & \textbf{59.0}  &  \textbf{77.2} \\
    ANP & 80.8 & 58.1 &  76.3 \\
    \bottomrule
  \end{tabular}
  \caption{ADD Results of CNP and ANP on Toy dataset.}
  \label{tab:ablation-aggregation}
\end{table}

\begin{figure}[htb!]
  \centering
    \begin{subfigure}{0.38\columnwidth}
    \includegraphics[width=\columnwidth]{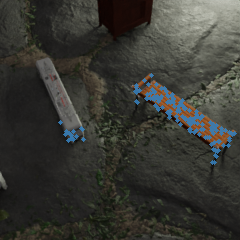}
    \caption{Feature ambiguity.}
    \label{fig:wrong-seg}
  \end{subfigure}
\ \ \ \ \ \ \ \ \ \ \ 
%\ \ 
%\hfill
    \begin{subfigure}{0.38\columnwidth}
    \includegraphics[width=\columnwidth]{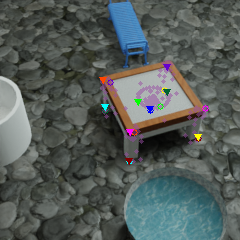}
    \caption{Symmetry ambiguity.}
    \label{fig:sym-kp}
  \end{subfigure}
  \caption{Limitations of the proposed method.}
  \label{fig:two-limitations}
\end{figure}

\textbf{Limitations.} We find two limitations of our method. First, we observe that in rare cases, our model suffers from \textit{Feature Ambiguity} by struggling to disentangle feature variations, \eg textures, shapes and lighting conditions. Sometimes it can be fooled by two similar objects which results in inaccurate segmentation (see \cref{fig:wrong-seg}). 
% More critically, the ambiguity of symmetric objects is not considered explicitly during training, which is still an open question and not well solved for keypoint-based approaches. 
Second, keypoint-based approaches suffer from \textit{Symmetry Ambiguity}, especially on novel objects where the symmetric axis is unknown. Consequently, keypoint predictions around the symmetric axis can be mismatched and hamper the training (see \cref{fig:sym-kp}).
% Considering a square-shaped table as an example, if the predicted keypoints are rotated \eg, 90 degree around the symmetry axis compared with the ground-truth, the estimated pose could be correct, however the training loss might be unreasonable high and thus misleads the model. 
% Examples for both limitations can be found in \cref{fig:two-limitations}. 
Accounting for the symmetry ambiguity, we also provide evaluations with the ADD-S metric in \cref{subsec:appendix-PBR} following prior work\newcite{ffb6d, DenseFusion, pvn3d}. 

% \vspace{-0.3cm}
\section{Conclusion}
In this paper, we present a CNP-based meta-learner for cross-category level 6D pose estimation, which is capable of extracting and transferring latent representation on unseen objects from only a few samples. Besides, we propose a simple yet effective geometry-aware keypoint detection module using GNN, which leverages the spatial connections between keypoints and improves generalization on unseen objects and robustness on occluded scenes. Furthermore, we create fully-annotated synthetic datasets called MCMS with various objects and categories, aiming to fill the vacancy for cross-category pose estimation. 
% We believe that our work can facilitate future research on generalized and transferable 6D pose estimation.

%%%%%%%%% References
%\clearpage
%\newpage
%\mbox{~}
\clearpage
{\small
\bibliographystyle{ieee_fullname}
\bibliography{References}
}

%%%%%%%%% Appendix
\newpage
%\include{latex/docs/6_appendix}
% use supplementary.tex
%\begin{appendices}
% \clearpage
\appendix
% reset the counter
\renewcommand{\thetable}{\arabic{table}}  
\renewcommand{\thefigure}{\arabic{figure}} 
\renewcommand{\theequation}{\arabic{equation}}
\setcounter{figure}{0}  
\setcounter{table}{0}  
\setcounter{equation}{0}

\section{Evaluation Results}
\label{sec:appendix-res}
We present evaluation results on both \linemod and MCMS datasets. More results and code are available at \url{https://github.com/Cvpr2022ID5164/CVPR2022-ID5164}.
 
%%%%%%%%%%%%%%%%%%%%%%%%%%%%%%%%%%%
\subsection{\linemod Dataset} 
\label{subsec:appendix-linemod}
\begin{figure*}[b]%
  \centering
    \includegraphics[width=\textwidth]{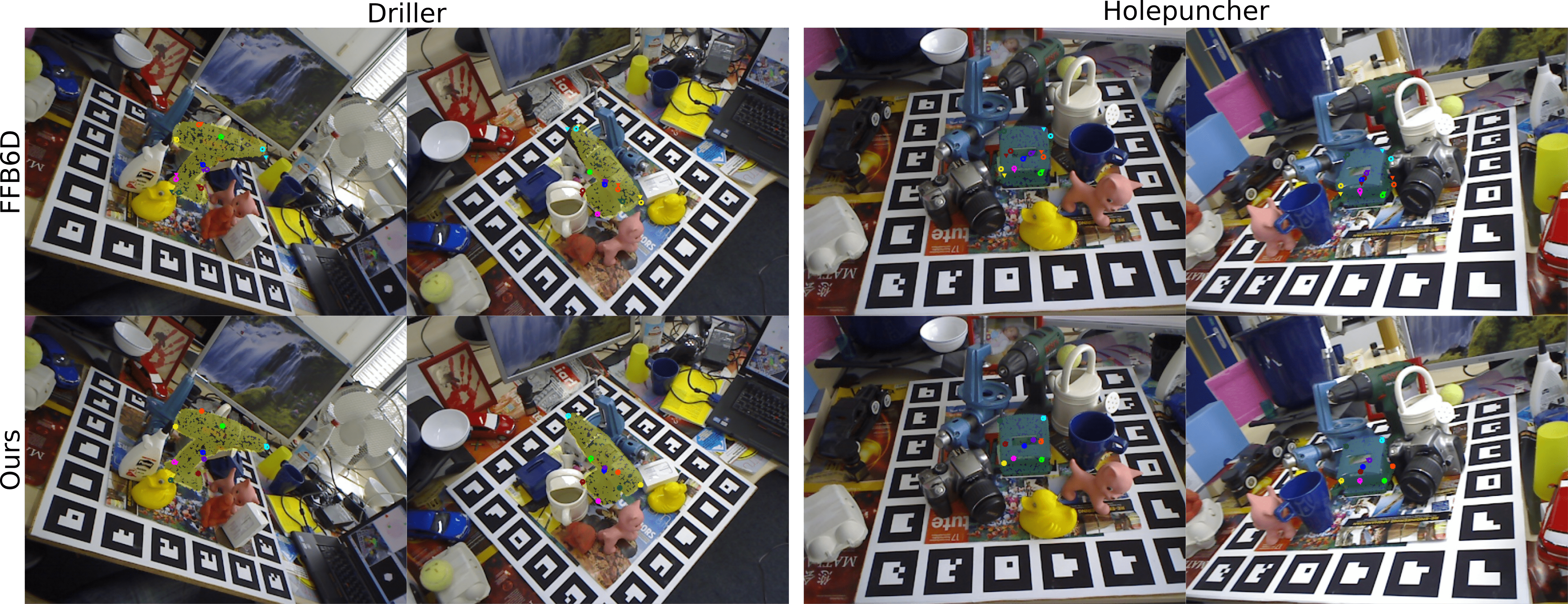}
    \caption{\textbf{Qualitative comparison on trained \linemod objects.} Triangles and circles are the projections of ground-truth and predicted keypoints respectively. It can be observed that keypoint predictions of our method are more accurate.}
    \label{fig:trained-linemod}
 \end{figure*}

\begin{figure*}[b]%
  \centering    
    \includegraphics[width=\textwidth]{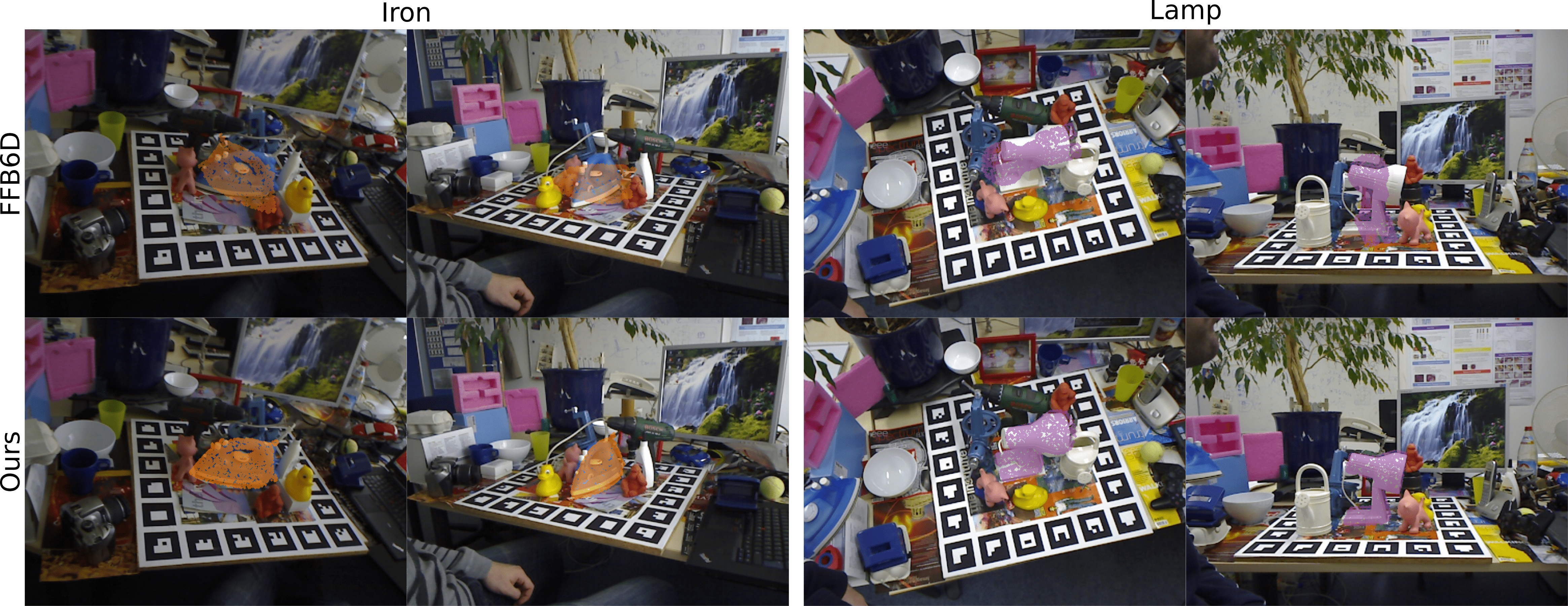}
    \caption{\textbf{Qualitative comparison on new \linemod objects.} Compared with FFB6D, the pose estimation on new objects of our GAML model is more accurate.}
    \label{fig:new-linemod}
\end{figure*}

%Qualitative Results 
The \linemod dataset\newcite{linemod-dataset}  is split into 10 training objects and 3 unseen test objects, where iron, lamp and phone are the novel test objects.  \cref{fig:trained-linemod} show the qualitative comparison between FFB6D\newcite{ffb6d} and the proposed model on training objects. It can be observed that our model can predict keypoints more accurately. From \cref{fig:new-linemod}, we can see that our model achieves better performance on novel objects. It should be noted that we only train one model for all objects, rather than train one model for each object respectively.

\comment{
\begin{table}[htb!]
    \centering
\begin{tabular}{@{}l|cc|cc@{}}\toprule
& \multicolumn{2}{c|}{FFB6D} & \multicolumn{2}{c}{Ours}
\\\midrule
%\\\cmidrule(lr){2-3}\cmidrule(lr){4-5}
Object & L1 Loss  & ADD-0.1d & L1 Loss  & ADD-0.1d      
\\\midrule
Ape & 0.09 & 100  & 0.019&100 \\
Holepuncher & 0.18 & 97.5 & 0.021 & 100  \\
\textbf{Iron}   & 1.53 & 0.0  & 0.26 &36  \\
\textbf{Lamp}  & 1.62 & 0.0 & 0.38 & 22 \\
\textbf{Phone} & 1.51 & 0.0  & 0.17 & 17  \\
\bottomrule
\end{tabular}
    \caption{\textbf{Evaluation results on LineMOD dataset.} Novel objects are marked as bold.}
    \label{tab:LineMod_checklater}
\end{table}
}

%%%%%%%%%%%%%%%%%%%%%%%%%%%%%%%%%%%
 \subsection{Toy-MCMS Dataset} \label{subsec:appendix-Toy}
\begin{figure*}[b]%
  \centering    
    \includegraphics[width=\textwidth]{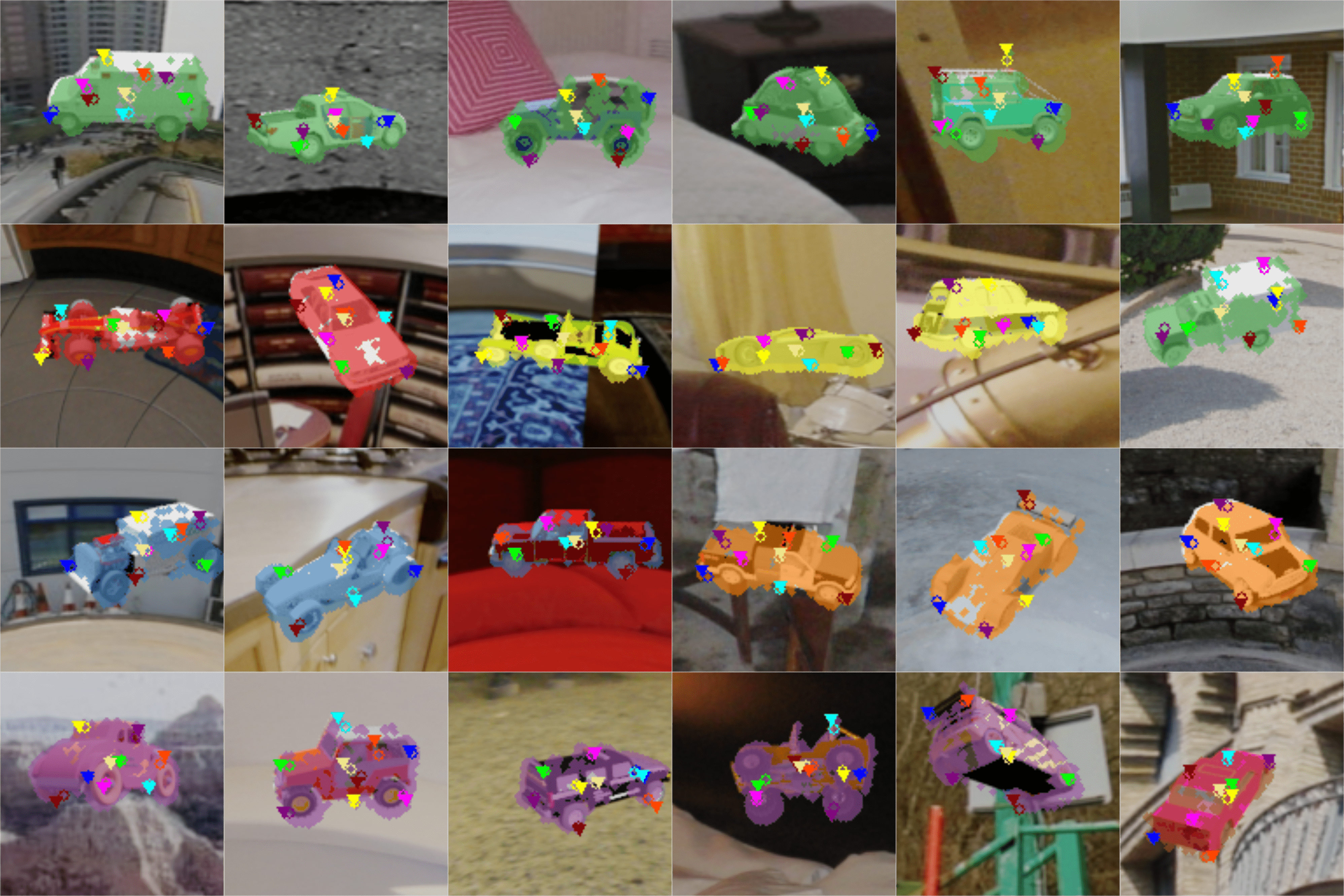}
    \caption{\textbf{Qualitative results on Toy-MCMS.} Our model can handle large intra-category variations. The car category is illustrated as an example.}
    \label{fig:toy-mcms-car}
\end{figure*}

\cref{tab:toy-single} provides the quantitative  results of inter-category 6D pose estimation on the car category. We use 50 images per object for training and vary the number of training objects. From the experimental results, 80 car objects can achieve a similar ADD accuracy as 1100 objects, while the training time is reduced evidently. Overall, this represents a good comprise between  prediction performance and training overhead. \cref{fig:toy-mcms-car} shows the qualitative results on novel test objects using the model trained with 80 objects. Note that even within the car category, the colors and shapes of novel objects still vary a lot.

 \begin{table}[H]
   \centering
   \begin{tabular}{@{}cccc@{}}
     \toprule
     \#Training objects & L1 loss[m] & ADD-0.1d[\%] & Time[h] \\
     \midrule
     1100 & 0.128 & 98.7  & 159\\
     80 &  0.245  & 96.7  &16\\
     \bottomrule
   \end{tabular}
   \caption{Single category - car evaluation on Toy-MCMS dataset}
   \label{tab:toy-single}
 \end{table}

%%%%%%%%%%%%%%%%%%%%%%%%%%%%%%%%%%%
%\subsection{Multi-category evaluation on Toy-MCMS dataset}\label{appendix:example}
\cref{tab:full-TOY-multi} shows the quantitative results of the
multi-category evaluation on the Toy-MCMS dataset.
The vanilla meta-learner (Vanilla-ML) using MLP decoder is compared with the proposed geometry-aware meta-learner (GAML). It is obvious that GAML outperforms the Vanilla-ML by a large margin. 

\begin{table}[htb!]
    \centering
\begin{tabular}{@{}l|cc|cc@{}}\toprule
 & \multicolumn{2}{c|}{Vanilla-ML}& \multicolumn{2}{c}{GAML}
\\\midrule
%\\\cmidrule(lr){2-3}\cmidrule(lr){4-5}
Category  & L1 Loss  & ADD-0.1d & L1 Loss  & ADD-0.1d      
\\\midrule
Airplane& 0.41 & 80.6 & 0.33 & \textbf{87.2}  \\
Bag & 0.34 & 85.2   & 0.30 & \textbf{87.1} \\
Basket& 0.83 & 49.7  & 0.62 & \textbf{65.1}  \\
Bathtub& 0.46 & 79.3  & 0.34 & \textbf{88.6}  \\
Bed & 0.72 & 60.1  & 0.57 & \textbf{72.0}  \\
Bench  & 0.95 & 56.7  & 0.63 & \textbf{72.3} \\
Birdhouse & 0.35 & 83.2  & 0.28 & \textbf{89.7}  \\
Bookshelf & 0.48 & 79.5 & 0.42 & \textbf{80.9}   \\
Cabinet & 0.39 & 83.5 & 0.31 & \textbf{89.7}   \\
Car & 0.31 & \textbf{92.9}  & 0.28 & \textbf{92.9}  \\
Camera& 0.57 & 65.0  & 0.46 & \textbf{73.4}   \\
Chair& 0.57 & 62.8   & 0.42 & \textbf{80.9}  \\
Helmet & 0.46 & 69.8 & 0.43 & \textbf{80.9}   \\
Motorcycle & 0.29 & 92.6  & 0.25 & \textbf{94.7}  \\
Mug & 0.25 & 91.9  & 0.24 & \textbf{93.3}   \\
Pillow & 0.76 & 54.7   & 0.58 & \textbf{81.5}  \\
Table & 0.80 & 61.1  & 0.54 & \textbf{76.2}  \\
Train & 0.41 & 80.6 & 0.36 & \textbf{86.2} \\
Vessel & 0.62 & 66.9 & 0.53 & \textbf{70.9} \\
Washer& 0.36 & 85.4 & 0.28 & \textbf{91.4}   \\
Bus* & 0.46 & 83.0& 0.42 & \textbf{85.4}   \\
Cap* & 0.71 & 46.6& 0.64 & \textbf{54.2}  \\
Laptop*& 1.02 & 18.8 & 0.73 & \textbf{48.8}   \\
Piano* & 0.86 & 47.1& 0.75 & \textbf{50.7}  \\
Remote* & 0.53 & 53.5& 0.52 & \textbf{56.1} \\
\midrule
Intra-Categ. & 0.52 & 74.2& 0.41 & \textbf{81.9} \\
Cross-Categ.  & 0.72& 50.3&  0.62 & \textbf{59.0}\\
All & 0.56 & 69.4& 0.45 & \textbf{77.2} \\
\bottomrule
\end{tabular}
    \caption{Multi-category evaluation on Toy-MCMS dataset}
    \label{tab:full-TOY-multi}
\end{table}

%%%%%%%%%%%%%%%%%%%%%%%%%%%%%%%%%%%
\subsection{PBR-MCMS Dataset}
\label{subsec:appendix-PBR}
We compare FFB6D, Vanilla-ML and GAML on intra- and cross-category levels. The full statistical summary can be found in \cref{tab:full-PBR-multi}. 
In general, the ADD metric is used for non-symmetric objects and ADD-S\newcite{linemod-dataset} for symmetric objects. Since the matching between points is ambiguous for some poses, ADD-S computes the mean distance based on the minimum point distance:
\begin{align} 
\mathrm{ADD-S}=\frac{1}{m}\sum_{x_1 \in \mathcal{O}} \min_{x_2\in\mathcal{O}}\norm{(Rx+t)- (R^{*}x+t^{*})}.
\end{align}\label{eq:adds-metric}

%%%%%%%%%%%%%%%%%%%%%%%%%%%%%%%%%%%5
\subsection{Occlusion-MCMS Dataset}\label{subsec:appendix-Occlusion}

Comparison between Vanilla-ML and GAML on Occlusion-MCMS is given in \cref{tab:full-occ-multi}.

\begin{table}[]
    \centering
\begin{tabular}{@{}l|cc|cc@{}}\toprule
& \multicolumn{2}{c|}{Vanilla-ML} & \multicolumn{2}{c}{GAML} 
\\\midrule
%\\\cmidrule(lr){2-3}\cmidrule(lr){4-5}
Category  & L1 Loss  & ADD-0.1d & L1 Loss  & ADD-0.1d      
\\\midrule
Airplane & 0.93 & 43.2  & 0.69 & \textbf{46.6} \\
Bag & 0.46 & 29.4 & 0.54 & \textbf{39.6}   \\
Bathtub  & 0.66 & 29.6 & 0.76 & \textbf{34.0} \\
Bed & 0.48 & \textbf{53.3}  & 0.71 & 44.2  \\
Bench & 0.66 & 40.4  & 0.67 & \textbf{49.0} \\
Bookshelf & 0.38 & 42.5 & 0.49 & \textbf{42.7 }  \\
Bus & 0.67 & 37.5 & 0.55 & \textbf{47.5}   \\
Cabinet & 0.48 & 35.2   & 0.44 & \textbf{55.2}  \\
Camera & 0.45 & 39.4 & 0.45 & \textbf{40.4}   \\
Cap & 0.39 & 42.1 & 0.45 & \textbf{43.1}   \\
Chair & 0.36 & 54.4  & 0.32 & \textbf{55.6}  \\
Earphone & 0.44 & 24.1 & 0.56 & \textbf{35.6}   \\
Motorcycle & 0.37 & \textbf{ 64.8} & 0.40 & 54.4   \\
Mug  & 0.38 & 40.2 & 0.34 & \textbf{52.9}   \\
Table & 0.62 & 23.3 & 0.72 & \textbf{29.4}   \\
Train & 0.52 & 32.3 & 0.53 & \textbf{36.2}   \\
Vessel  & 0.60 & \textbf{40.3} & 0.90 & 33.4  \\
Washer & 0.46 & 37.7  & 0.44 & \textbf{55.0} \\
Printer & 0.56 & 27.5 & 0.57 & \textbf{39.6}   \\
Birdhouse* & 0.44  & 23.5 & 0.51 &  \textbf{28.0} \\
Car* & 0.52 & 42.9 & 0.50 & \textbf{44.4} \\
Laptop* & 0.47 & 26.0& 0.41 & \textbf{47.7}   \\
Piano* & 0.51 & 27.3& 0.66 & \textbf{32.5}  \\
Sofa* & 0.48 & 45.4 & 0.44 & \textbf{57.9} \\
\midrule
Intra-Categ.  & 0.52 & 38.7 & 0.55 & \textbf{43.9}\\
Cross-Categ.  & 0.48 & 33.0 & 0.51 & \textbf{42.1} \\
All & 0.51 & 37.6 & 0.54 & \textbf{43.5} \\
\bottomrule
\end{tabular}
    \caption{Multi-category evaluation on Occlusion-MCMS dataset}
    \label{tab:full-occ-multi}
\end{table}

%%%%%%%%%%%%%%%%%%%%%%%%%%%%%%%%%%%
\section{Ablation Study} \label{sec:appendix-ablation}

%%%%%%%%%%%%%%%%%%%%%%%%%%%%%%%%%%%
\subsection{Effect of K Neighbors in GNN}\label{subsec:appendix-ablation-k}
We measure the ADD-0.1d accuracy of multi-category and single-category training with $k=3$ and $k=8$ in the GNN decoder.  \cref{tab:full-ablation-knn} presents the quantitative results. 

\begin{table}[t]
  \centering
  \begin{tabular}{@{}c|c|cc@{}}
    \toprule
     &  & k = 3 & k = 8\\
    \midrule
    %Intra-Categ. & 44.9 & 62.9  \\
    %Inter-Categ. &  42.4 & 58.6 \\
    %All &  44.4  & 62.0 \\
    \multirow{3}{*}{Multi.}& Intra-Categ.  &60.1 & \textbf{62.9}  \\
    &Cross-Categ.  & \textbf{60.2} & 58.3 \\
    &All &  60.1   &\textbf{61.9} \\  
        \midrule
    \multirow{13}{*}{Single.}&Airplane &89.1& \textbf{93.4} \\
    %Bag(t.b.d.)  & \textbf{53.8} & 53.4 \\
    %Birdhouse(delete this!)& \textbf{62.2} & 60.6 \\
    &Bathtub & 43.8 & \textbf{46.9}\\
    &Bench & 74.0 & \textbf{75.6}\\
    %Bus &23.8 &\textbf{75.0} \\
    &Camera &  67.0 & \textbf{69.6} \\
    &Car &  90.6 & \textbf{94.6} \\
    &Chair  & \textbf{93.3} & 88.6 \\
    &Laptop  & \textbf{99.6} & 99.1 \\
    &Motorcycle & \textbf{94.4} & \textbf{94.4} \\
    &Piano &  74.7 & \textbf{98.1} \\
    &Sofa &  91.2 & \textbf{96.6} \\
    &Vessel  &  45.7 & \textbf{69.0}\\
    &Washer & 72.7 & \textbf{73.5} \\
    &Mean & 77.9 & \textbf{83.3}\\
    \bottomrule
  \end{tabular}
  \caption{Effect of $K$ neighbors in GNN decoder on PBR-MCMS. The first block provides the statistic for multi-category training, where one model is trained on multiple categories and tested on new objects of both training and new categories. For single category training, each model is trained and tested per category. }
  \label{tab:full-ablation-knn}
\end{table}

%%%%%%%%%%%%%%%%%%%%%%%%%%%%%%%%%%%5
\subsection{Effect of Aggregation Module in CNP}\label{subsec:appendix-ablation-attention}
\begin{figure}[]
\centering
	\includegraphics[width=\columnwidth]{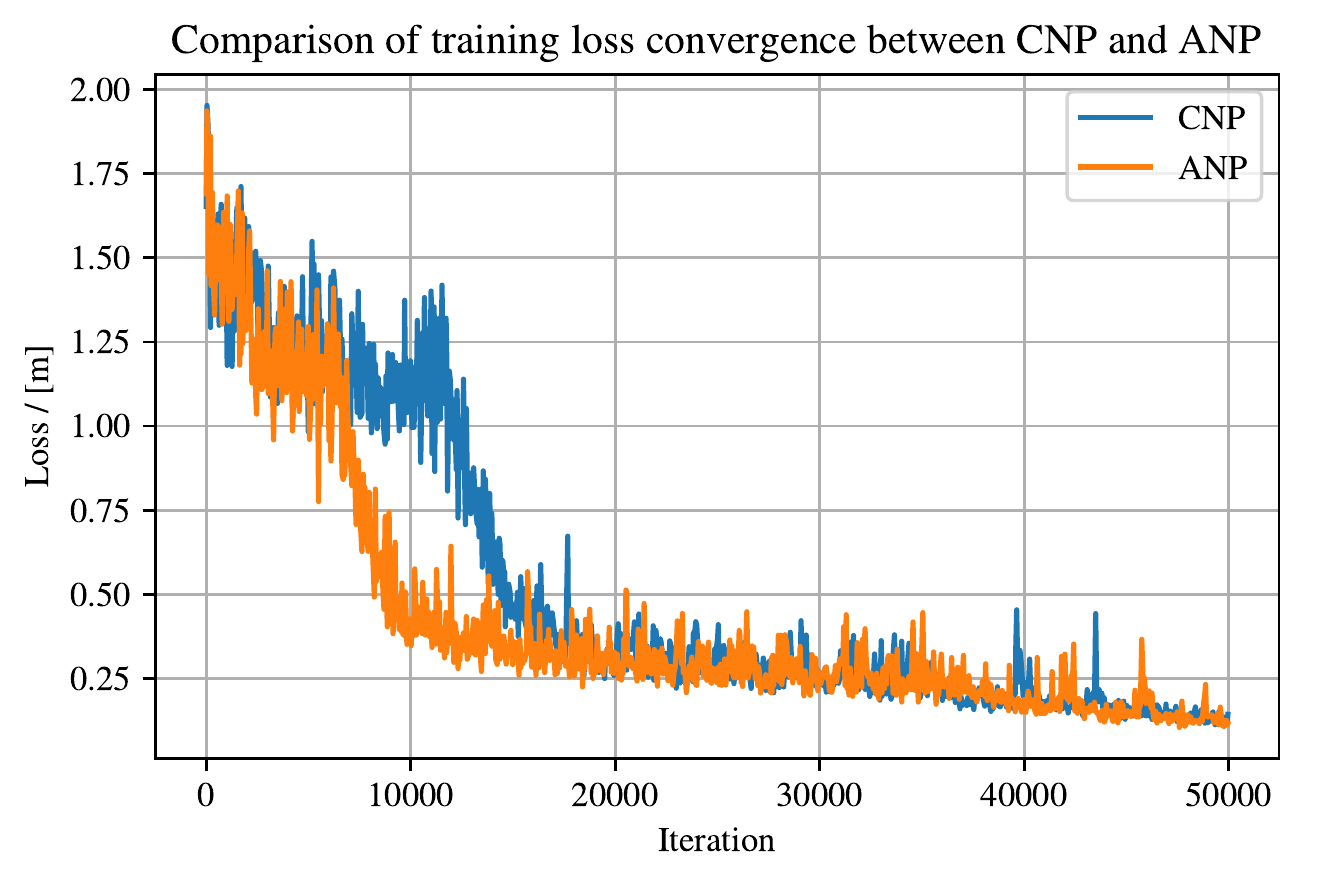}
		\caption[]{Comparison of training loss between CNP and ANP}
		\label{fig:ablation-aggregation-app} 
\end{figure}

\cref{fig:ablation-aggregation-app} shows  that  both CNP and ANP methods  achieve  similar  training  performance in the end, even though ANP converges faster at the beginning.

%%%%%%%%%%%%%%%%%%%%%%%%%%%%%%%%%%%
\section{Network Architecture} \label{sec:implementation-details}
%\subsection{Limitations}\label{subsec:appendix-limitations}
The detailed architecture model is shown in \cref{tab:architecture}. We use ReLU as activation function after each FC layer except the output layer of segmentation decoder and global GNN decoder for keypoint offset prediction.

%%%%%%%%%%%%%%%%%%%%%%%%%%%%%%%%%%%
%\section{Limitations} \label{sec:appendix-limitations}
%\subsection{Limitations}\label{subsec:appendix-limitations}
% \comment{
% \cref{fig:limitations} presents the failure cases. In \cref{fig:wrong-seg}, the target is misclassified due to the appearance of another object with similar shape. \cref{fig:sym-kp} illustrates the ambiguity of symmetric objects.T  The predicted keypoints (circles) are rotated 90 degree around the symmetry axis compared with the ground-truth (triangles), 
% which results in quite accurate pose estimation, nevertheless gives an unreasonable high training loss and misleads the model. 

% \begin{figure}[htb!]
%   \centering
%     \begin{subfigure}{0.45\columnwidth}
%     \includegraphics[width=\columnwidth]{latex/figs/wrong_seg.png}
%     \caption{Inaccurate segmentation}
%     \label{fig:wrong-seg}
%   \end{subfigure}
%   %\hfill
%     \begin{subfigure}{0.45\columnwidth}
%     \includegraphics[width=\columnwidth]{latex/figs/symmetric_table.png}
%     \caption{Ambiguity of symmetric objects}
%     \label{fig:sym-kp}
%   \end{subfigure}
%   \caption{Limitations}
%   \label{fig:limitations}
% \end{figure}
% }

\begin{table}[]
  \centering
  \begin{tabular}{@{}c|c|c@{}}
    \toprule
    Component & Layer & Output Size\\
    \midrule
    \multirow{3}{*}{Encoder}& FC & 128 \\
    & FC & 128 \\
    & FC &  128 \\  
        \midrule
    \multirow{4}{*}{Seg. Decoder} & FC & 128\\
    & FC &128\\
    & FC &128\\
    & FC &2\\
        \midrule
    \multirow{3}{*}{Local GNN} & FC & 128\\
    & FC &128\\
    & FC &128\\
        \midrule
    \multirow{3}{*}{Global GNN} & FC & 128\\
    & FC &128\\
    & FC & 3 \\
    \bottomrule
  \end{tabular}
  \caption{GAML network architecture.}
  \label{tab:architecture}
\end{table}

%%%%%%%%%%%%%%%%%%%%%%%%%%%%%%%%%%%%
\begin{table*}[]
    \centering
\begin{tabular}{@{}l|ccc|ccc|ccc@{}}\toprule
& \multicolumn{3}{c|}{FFB6D} & \multicolumn{3}{c|}{Vanilla-ML} &\multicolumn{3}{c}{GAML} 
\\\midrule
%\\\cmidrule(lr){2-3}\cmidrule(lr){4-5}
Category  & L1 Loss  & ADD & ADD-S &L1 Loss  & ADD & ADD-S & L1 Loss  & ADD & ADD-S
\\\midrule
Airplane & 1.51 & 9.1 & 85.7 &0.11 & \textbf{90.4}& 96.8 & 0.11 & 89.8 & 98.8 \\
Bag & 1.98 & 5.1 &48.1 &0.41 & 40.0 & 85.0 & 0.47 & \textbf{42.7}& 87.1  \\
Bathtub & 2.22 & 2.7 & 41.4 &0.55 &43.3 & 86.7 & 0.60 & \textbf{45.2}& 90.8 \\
Bed  & 2.31 & 2.9 & 33.3 & 0.31 & \textbf{72.3} & 90.4  &0.41 & 58.5 & 90.8\\
Bench  & 2.26 & 2.9 & 43.0 &0.39 &62.1 &91.4 & 0.35 & \textbf{69.8} &91.7\\
Bookshelf& 2.28 & 2.4  & 32.6  &0.36 & \textbf{55.0} & 85.4 & 0.41 & 50.2&77.9 \\
Bus & 1.94 & 3.5  & 56.5 &0.51 &41.5 &89.6  & 0.36 & \textbf{69.8}& 92.7 \\
Cabinet & 2.38 & 2.2  & 24.0 &0.43 & 53.5 & 73.7 & 0.34 & \textbf{67.7} & 83.5 \\
Camera & 1.93 & 2.1  &51.5 &0.38 & 46.3 &86.3  & 0.34 & \textbf{54.8} &85.8 \\
Cap &1.75 & 3.1  & 68.5 &0.19 &79.2 & 98.5 & 0.19 & \textbf{80.8}&98.8 \\
Chair & 2.27 & 1.1 & 26.1  &0.18 &\textbf{80.0} &93.1 & 0.19 & \textbf{80.0}&89.6 \\
Earphone & 1.79 & 4.2  & 62.8 &0.38 & 34.0 & 86.2 & 0.43 & \textbf{49.2} &97.1 \\
Motorcycle & 1.65 & 12.7  & 85.1 &0.16 &\textbf{90.2} & 98.5 & 0.21 & 85.6&98.1 \\
Mug &2.08 & 0.9 &43.7 &0.12 &\textbf{86.8} & 97.9 & 0.14 & 84.2 & 94.4   \\
Table &2.38 & 2.0  & 19.2 &0.61 &33.1 & 73.5 & 0.65 & \textbf{39.2} &93.1 \\
Train &1.67 & 12.9  & 71.9 &0.46 &38.5 &85.8  & 0.49 & \textbf{47.7} &90.6 \\
Vessel &1.64 & 11.0  & 68.9 &0.35 &\textbf{57.7}&94.1 & 0.37 & 56.0 & 90.4 \\
Washer  &2.48 & 4.1 &29.1 &0.33 & 54.8 & 85.4 & 0.30 & \textbf{68.1} &89.0 \\
Printer &2.21 & 1.0 &33.1 &0.41 &47.9 &  83.9 & 0.43 & \textbf{55.2} &80.0  \\
Birdhouse*  & 2.09 & 0.8 & 21.0 &0.39 & \textbf{35.6} & 59.4 &0.43 &  35.4 &64.6 \\
Car* &1.79 & 2.4 & 70.5 & 0.44 & 52.5 & 97.1  & 0.43 & \textbf{56.9} &96.7 \\
Laptop* &2.22 & 1.3  &10.5 & 0.32 & 54.0 &82.9 & 0.20 & \textbf{85.0} &93.1 \\
Piano* &2.04 & 2.0 &39.3 &0.43 &\textbf{45.8}&77.5 &0.44 & \textbf{45.8} &80.0 \\
Sofa*  &2.17 & 2.6 & 27.1& 0.34 & 68.1 &84.6  & 0.34 & \textbf{69.8} &79.0\\
\midrule
Intra-Categ.& 2.03 & 4.53 & 48.7 & 0.35 & 58.2& 88.5  & 0.36 & \textbf{62.9}&90.0 \\
Cross-Categ. & 2.06 & 1.81 &  33.7 & 0.38 & 51.2 & 80.2  & 0.37 & \textbf{58.6}&82.7 \\
All & 2.04 & 3.96 & 45.5 & 0.36 &56.7 & 86.8 & 0.36 & \textbf{62.0}&88.4 \\
\bottomrule
\end{tabular}
    \caption{Multi-category evaluation on PBR-MCMS dataset}
    \label{tab:full-PBR-multi}
\end{table*}

%%%%%%%%%%%%%%%%%%%%%%%%%%
 %\vfill 

\begin{figure*}[b]%
\centering
\includegraphics[width=\textwidth]{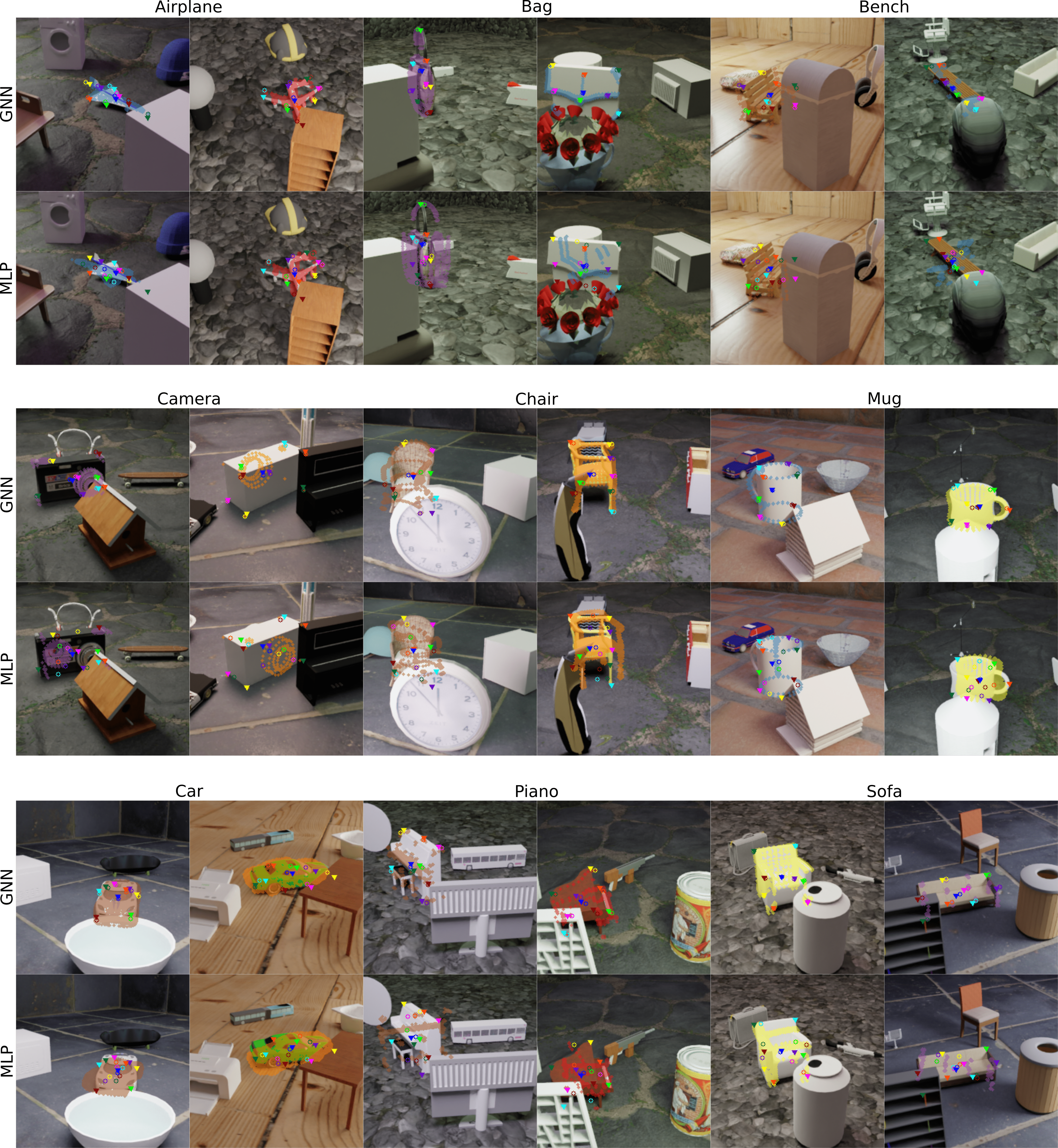}
\caption[]{\textbf{Qualitative comparison between GNN and MLP decoder on Occlusion-MCMS.} Triangles and circles are the projected ground-truth and predicted keypoints respectively.}
\label{fig:gnn_vs_mlp_occ}
\end{figure*}

\end{document}